\documentclass[11pt]{article} 
\usepackage[]{acl}

\usepackage{times}
\usepackage{latexsym}
\usepackage{subcaption}
\usepackage{amsmath, amssymb}

\usepackage[T1]{fontenc}
\usepackage[utf8]{inputenc}

\usepackage{amsmath,amsfonts,bm}

\def\eqref#1{equation~\ref{#1}}

\def\1{\bm{1}}

\DeclareMathAlphabet{\mathsfit}{\encodingdefault}{\sfdefault}{m}{sl}
\SetMathAlphabet{\mathsfit}{bold}{\encodingdefault}{\sfdefault}{bx}{n}

\usepackage{hyperref}
\usepackage{url}
\usepackage{enumitem} 

\usepackage{xcolor}      
\usepackage{algorithm}
\usepackage{algpseudocode} 
\usepackage{booktabs,multirow,tabularx}

\usepackage{graphicx}   
\usepackage{arydshln}  
\usepackage{makecell}
\usepackage{microtype}
\usepackage{subcaption}
\usepackage{setspace} 

\usepackage{xcolor}
\usepackage{pifont}  

\usepackage[most]{tcolorbox}
\tcbset{
  colback=black!2,
  colframe=black!35,
  boxrule=0.4pt,
  arc=2pt,
  left=8pt,right=8pt,top=6pt,bottom=6pt,
  fonttitle=\bfseries,
  before skip=10pt, after skip=10pt
}

\title{Value of Information: A Framework for Human–Agent Communication}
\author{
Yijiang River Dong\textsuperscript{1},
Tiancheng Hu\textsuperscript{1},
Zheng Hui\textsuperscript{1},
Caiqi Zhang\textsuperscript{1}\\
\textbf{Ivan Vuli\'c}\textsuperscript{1},
\textbf{Andreea Bobu}\textsuperscript{2}\thanks{ Equal advising.},
\textbf{Nigel Collier}\textsuperscript{1}\footnotemark[1] \\
\textsuperscript{1}University of Cambridge \quad
\textsuperscript{2}MIT \\
\texttt{\{yd358,th656,zh403,cz391,iv250,nhc30\}@cam.ac.uk}\\
\texttt{abobu@mit.edu}
}

\begin{document}
\maketitle
\begin{abstract}
Large Language Model (LLM) agents deployed for real-world tasks face a fundamental dilemma: user requests are underspecified, yet agents must decide whether to act on incomplete information or interrupt users for clarification. Existing approaches either rely on brittle confidence thresholds that require task-specific tuning, or fail to account for the varying stakes of different decisions. We introduce a decision-theoretic framework that resolves this trade-off through the Value of Information (VoI), enabling agents to dynamically weigh the expected utility gain from asking questions against the cognitive cost imposed on users. Our inference-time method requires no hyperparameter tuning and adapts seamlessly across contexts—from casual games to medical diagnosis. Experiments across four diverse domains (20 Questions, medical diagnosis, flight booking, and e-commerce) show that VoI consistently matches or exceeds the best manually-tuned baselines, achieving up to 1.36 utility points higher in high-cost settings. This work provides a parameter-free framework for adaptive agent communication that explicitly balances task risk, query ambiguity, and user effort. Our code will be available at \url{https://github.com/dong-river/VOI_communication}.
\end{abstract}


\section{Introduction} 
LLM agents are increasingly deployed as autonomous collaborators in complex, real-world tasks. However, a fundamental bottleneck remains: user requests are inherently underspecified, carrying latent goals, contexts, and unstated preferences \cite{malaviya_contextualized_2024, yao2024tau, peng2024plga, dong2024can, hui2025safe}. A request to ``book a flight to London'' omits critical details, such as budget constraints, preferred departure times, tolerance for layovers. No amount of model capability can resolve this ambiguity without external input; the agent must ask. Yet excessive questioning frustrates users and undermines the agent's value proposition. Effective collaboration thus requires agents to balance two risks: acting on incomplete information and misaligning with user intent, or interrupting frequently and imposing cognitive burden.

Current approaches fall short in navigating this trade-off. Fixed-round strategies ask a predetermined number of questions regardless of context, ignoring task-specific needs. Adaptive methods trigger clarification when model confidence falls below a manually-tuned threshold, but this threshold selection is brittle and fails to generalize across domains or cost structures. Neither approach explicitly reasons about whether the information gained justifies the user's effort.

We argue that agents should treat communication as a rational decision, asking questions only when the expected improvement in task outcomes justifies the user's time and effort. We adopt a Rational Speech Act (RSA) perspective \cite{goodman2016pragmatic, frank2012predicting} viewing dialogue as a rational action. Building on prior RSA work on interactive questioning-answering \cite{hawkins2015you} and utility-grounded pragmatic reasoning \cite{sumers2021extending}, the agent should only ask questions when the expected benefit of improved downstream decisions outweighs the cost of additional interaction—capturing both cost of communication \cite{hawkins2015you} and utility of downstream decisions \cite{sumers2021extending}. Under this lens, we formalize the clarify-or-commit decision through three contextual factors: (1) \textbf{Query Ambiguity}: the degree of uncertainty about the user's true intent; (2) \textbf{Task Risk}: the severity of the consequences of a wrong action; and (3) \textbf{Cognitive Load}: the cost, in time and effort, imposed on the user by asking for clarification. 

To operationalize this reasoning, we propose a decision-theoretic framework grounded in the Value of Information (VoI), a classic principle from decision theory \cite{raiffa1961applied}. Our inference-time method allows an LLM to explicitly calculate the expected utility gain of asking a potential question, weighing it directly against the communication cost. This provides a principled mechanism for the agent to decide whether the information it might receive is worth the user's attention. Our contributions are threefold: \textbf{(a)} We formalize the adaptive communication problem in human-agent interaction from a decision-theoretic perspective, identifying three key factors: ambiguity, risk, and cognitive load.
\textbf{(b)} We propose a practical, inference-time VOI-based method that allows an LLM to estimate these contextual factors and dynamically decide whether to act or to seek clarifications \textbf{(c)} We demonstrate through experiments across four distinct domains: 20 Questions, medical diagnosis, flight booking, and online shopping, that our parameter-free VoI method automatically identifies the optimal operating point. Across varying communication costs, VoI matches or exceeds the best manually-tuned baselines in 18 of 20 conditions, achieving utility gains of up to 1.36 points in high-cost settings.

\begin{figure*}[!t]
    \includegraphics[width=\linewidth]{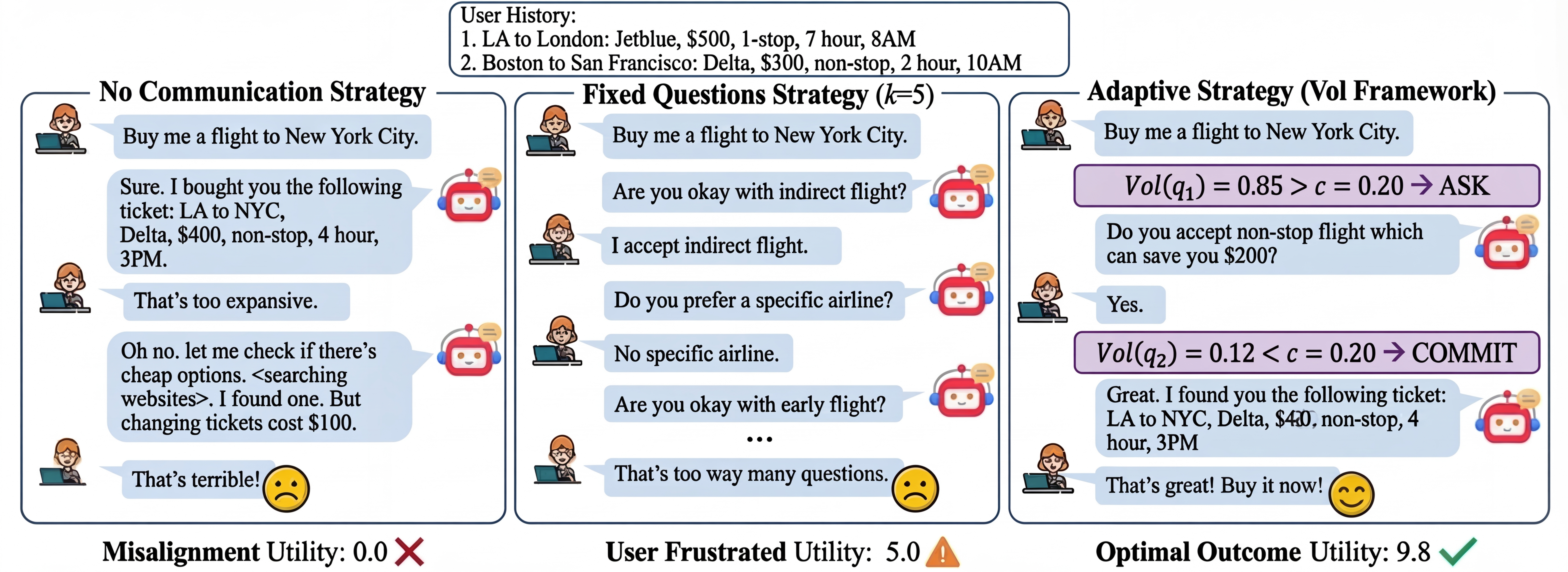}
    \caption{\textbf{Illustration of different communication methods and user reaction.} Given user flight history, an LLM agent is able to infer user latent preferences with some probability. Excessive questions that asks about every aspect of preference would lead to user dissatisfaction (A) while directly acting without communication could lead to unexpected consequences (B). Decision-theoretic reasoning can balance expected utility gain via asking user questions against communication cost to achieve efficient but effective communication at inference time (C).} 
    \label{fig:communication_methods}

\end{figure*}

\section{Related Work}
\paragraph{Standard LLM Agent Paradigm.} Our work is situated within the broader context of developing autonomous LLM agents. Much foundational research in this area focuses on improving agent reasoning, planning, and tool-use capabilities. Prominent paradigms like \citet{yao2023react} and others are often evaluated in benchmarks that, while complex, assume the user's initial instruction is complete and unambiguous~\citep{yao2022webshop,zhou2023webarena,xie2024osworld}. This focus on task execution rather than the real-world productivity users expect from agents, leaving a critical gap for truly deploying agents \cite{sun2025training,  shah2024agents, zhou2025usereffective, hui-etal-2025-winspot}.

Recently, a new wave of research has begun to address agent reliability by introducing principled frameworks from decision theory~\citep{liu2024dellma,lin-etal-2024-decision,chen2025decisionflow}. However, these approaches typically focus on making an optimal decision given a static, pre-defined state of information. Our work bridges these two areas: we adopt the rigor of decision theory but focus on the upstream problem of active information gathering, allowing the agent to dynamically resolve ambiguity before committing to an action.

\paragraph{LLM Proactive Communication.} Prior work has explored prompting techniques to improve LLM interactivity. These methods can elicit user preferences~\citep{li_eliciting_2023} or encourage active disambiguation of ambiguous queries~\citep{deng2023prompting, zhang2024ask}.  While prompting can directly induce clarifying behaviors, prior work shows that the resulting strategies are often suboptimal without more principled planning or learning algorithms. Our work provides such a principled algorithm to govern the agent's communication decisions.

\paragraph{Uncertainty-Gated and Information-Theoretic Methods.} A more systematic approach uses model-uncertainty estimates to decide when to seek clarification, triggering a question when prediction confidence or entropy falls below a selected threshold~\citep{wang2025adaptive, zhang2023clarify, kuhn_clam_2023, ren2023robots, grand2025shoot}. While an improvement over heuristics, these information-centric views can be insufficient, as they do not directly consider the downstream task's stakes. Our method addresses this by employing the Value of Information (VoI)~\citep{raiffa1961applied,4082064}, a core concept from decision theory. Instead of measuring information gain in isolation, VoI measures how that information is expected to improve the utility of the final action, explicitly connecting the purpose of communication to the stakes of the decision. 

\paragraph{Learning-Based Approaches.} Different from the inference-time algorithms above, another line of research uses reinforcement learning to improve LLM collaboration with humans. Variants of Direct Preference Optimization (DPO) have been applied to encourage models to request clarification when needed~\citep{zhang_modeling_2024, chen_learning_2024, wu2025collabllm, qian2025userrl, sun2025training}. However, RL is often task-specific, requiring a carefully designed simulation environment and training pipeline, which is fundamentally different from our VOI-based method which operate purely at inference-time.

\paragraph{Rational Speech Act} RSA-style pragmatic models cast language as (approximately) rational action: speakers choose utterances to shape a listener’s inferences under explicit priors and costs \cite{frank2012predicting,goodman2016pragmatic}. Beyond single-shot reference, RSA has been extended to interactive question–answering, where questions are selected to trade off expected informativeness against asking cost \cite{hawkins2015you}, and to action-oriented settings where the point of communication is not only belief change but improving downstream decisions (e.g., signaling bandits) \cite{sumers2021extending}. Researchers then extend to “Neural RSA” that replace hand-specified literal models with learned speakers/listeners in grounded tasks \cite{andreas-klein-2016-reasoning, monroe-etal-2017-colors}. Most recently, RSA has been adapted to the era of LLMs, serving both as an inference-time control to guide generation \cite{wang-demberg-2024-rsa, cao2025pragmaticreasoningimprovesllm}.

\section{Problem Formulation}
We formulate the adaptive communication task as a sequential decision-making process where an LLM agent interacts with a user to select an optimal action.

\paragraph{Preliminaries.}
The agent receives an initial, potentially ambiguous, user query~$S$. The user's true goals and preferences are represented by a latent state $\theta \in \Theta$, which is not directly observable by the agent. The agent has access to a set of possible terminal actions $a \in \mathcal{A}$. To resolve ambiguity about $\theta$ and choose the best action $a^*$, the agent can engage in a multi-turn dialogue with the user.

\paragraph{The Clarify-or-Commit Process.}
The interaction proceeds in a sequence of turns. At each turn $t$, given the dialogue history $H_t = (q_1, u_1, \dots, q_{t-1}, u_{t-1})$, the agent must make a decision:
\begin{enumerate}
    \item \textbf{CLARIFY:} Select and pose a question $q_t$ from a set of possible questions $\mathcal{Q}$. Upon receiving the user's answer $u_t$, the history is updated to $H_{t+1}$ and the process continues.
    \item \textbf{COMMIT:} Terminate the dialogue and select a final action $a \in \mathcal{A}$ based on the current history $H_t$.
\end{enumerate}
The agent's strategy for making this choice at each turn is the \textbf{clarify-or-commit} policy, which is the central object of our study. This simple clarify-or-commit choice lies at the heart of adaptive communication: every question carries both the potential to reduce uncertainty and the cost of additional user effort.

\paragraph{Utility and Objective.}
The success of a committed action $a$ is measured by a utility function $U(\theta, a)$, which quantifies how well the action aligns with the user's true latent state $\theta$. Communication incurs a cost $c(H)$, representing the user's cognitive load, which quantifies the time and effort user spent on the dialogue. If the agent commits to action $a$ after a final history $H$, the total utility is $U(\theta, a) - c(H)$. The agent's objective is to devise a policy that maximizes the expected total reward, optimally balancing the utility gain from asking questions against cumulative communication cost.

\section{Methods}

To address the clarify-or-commit problem, an agent requires a principled policy for deciding when the potential benefit of asking a question outweighs the cost of interaction. Simple heuristic-based strategies often fail because they do not explicitly reason about the downstream consequences or the stakes of the decision. To overcome this limitation, we propose an adaptive policy grounded in the Value of Information (VoI), a core concept from decision theory \cite{raiffa1961applied}.

\subsection{Value of Information Framework}
\label{sec:voi_theory}
The baselines above are either non-adaptive or rely on generic, task-agnostic heuristics like confidence. They fail to explicitly reason about the \textit{value} of the information a question might provide in the context of heterogeneous task stakes and unequal feature importance. To address this, we formalize our approach using the VoI framework.

\paragraph{Beliefs and Expected Utility.}
Let $\Theta$ be the set of possible latent user intents (e.g., the specific product features preferred or the true medical condition). The agent maintains a belief distribution $b(\theta)$ over $\Theta$. Given this belief, the expected utility (EU) of committing to a terminal action $a \in \mathcal{A}$ is:
\begin{equation}
    \text{EU}(a \mid b) = \mathbb{E}_{\theta \sim b}[U(\theta, a)] = \sum_{\theta \in \Theta} b(\theta) U(\theta, a).
    \label{eq:expected_utility}
\end{equation}
If the agent were to commit immediately, it would choose the action $a^* = \arg\max_{a \in \mathcal{A}} \text{EU}(a \mid b)$. The utility of this decision is the value of acting under the current belief $b$:
\begin{equation}
    V(b) = \max_{a \in \mathcal{A}} \text{EU}(a \mid b).
    \label{eq:value_function}
\end{equation}

\paragraph{Calculating the Value of a Question.}
To evaluate a potential question $q$, the agent considers the set of possible answers $\mathcal{Y}$. For any given answer $y \in \mathcal{Y}$, the agent would update its belief to a posterior $b_y(\theta) = P(\theta \mid H, q, y)$. The expected value of the decision \textit{after} receiving an answer to question $q$ is the expectation over all possible answers $y$:
\begin{equation}
    V_{\text{post}}(b, q) = \sum_{y \in \mathcal{Y}} p(y \mid q, b) \cdot V(b_y),
    \label{eq:posterior_value}
\end{equation}
where $p(y \mid q, b)$ is the probability of receiving answer $y$ given the current belief. In practice, to make  computation feasible, we restrict the answer space to a closed set of multiple choice or yes-no questions. For each sampled hypothesis $\theta$, we query the LLM to simulate the likelihood of each response $y$ given question $q$, aggregating these to find the marginal probability $p(y \mid q, b)$.

The \textbf{Value of Information} for question $q$ is the difference between the expected utility after asking and the utility of acting now:
\begin{equation}
    \text{VoI}(q) = V_{\text{post}}(b, q) - V(b).
    \label{eq:voi}
\end{equation}

\paragraph{The Clarify-or-Commit Policy.}
Our framework uses this VoI calculation to establish a decision rule. At each turn, the agent evaluates the net utility gain for each candidate question:
\begin{equation}
    \text{NetVoI}(q) = \text{VoI}(q) - c,
    \label{eq:net_voi}
\end{equation}
where $c$ is the per-question communication cost. The agent selects the question $q^*$ with the highest positive net value. If $\max_q \text{NetVoI}(q) \le 0$, the expected utility gain from further communication is not worth the cost. The agent terminates the dialogue and commits to the best action under its current belief.

\subsection{Instantiation with LLMs}
\label{sec:voi_impl}
While Section \ref{sec:voi_theory} establishes the theoretical foundations of our approach, in this section, we describe how we leverage LLMs to approximate these components at inference time.

\paragraph{Estimating and Updating Belief Distributions.}
Given the set of candidate latent factors $\Theta$, we prompt the LLM to explicitly quantify its uncertainty by outputting a probability distribution $b(\theta)$ over these factors. Different from standard Bayesian approaches update beliefs analytically via a fixed likelihood function, we employ a LLM to estimate the probability distribution over $\Theta$ \cite{liu2024dellma,kobalczyk2025active,hu2025simbench,chen2026decoupling}. To obtain the posterior belief $b_y$ required for Eq.~\ref{eq:posterior_value}, we feed the history augmented with a simulated interaction (question $q$ and hypothetical answer $y$) back into the model and prompt it to re-estimate the distribution over $\Theta$. This allows the agent to dynamically update its confidence based on the semantic content of the answer.

\paragraph{Simulating User Responses.}
To calculate the expected value of a question, we perform a one-step lookahead simulation \cite{kobalczyk2025active} to estimate the marginal likelihood of possible answers $p(y \mid q, b)$. To ensure computational tractability in Eq.~\ref{eq:posterior_value}, we constrain the agent to ask closed-ended questions (e.g., multiple-choice or Yes-No questions), thereby defining a finite answer space $\mathcal{Y}$. The probability of each response is computed by marginalizing over the current beliefs: $p(y \mid q, b) \approx \sum_{\theta \in \Theta} p(y \mid q, \theta) b(\theta)$, where the term $p(y \mid q, \theta)$ represents the LLM's prediction of the user's response assuming $\theta$ is the ground truth.

\begin{algorithm*}[t]
\scriptsize
\caption{VOI Algorithm}
\label{alg:voi}
\begin{algorithmic}[1]
\Require Instruction $S$; action set $\mathcal{A}$; utility $U(\theta,a)$; question generator $\mathrm{GenQ}$; belief updater $\mathrm{Update}$; cost $c(\cdot)$; clarification budget $K_{\max}$
\State $H \gets \{S\}$;\quad $b \gets \mathrm{Prior}(S)$
\For{$t = 1,2,\dots,K_{\max}$}
    \State $Q \gets \mathrm{GenQ}(H)$ \Comment{small set of targeted questions}
    \State $V_0 \gets V(b) = \max_{a \in \mathcal{A}} \mathbb{E}_{\theta \sim b}[U(\theta,a)]$
    \ForAll{$q \in Q$}
        \State Sample plausible replies $\{(y_k,\pi_k)\}_{k=1}^{K}$ from $P(\cdot \mid b,q)$
        \State $V_q \gets \sum_{k=1}^{K} \pi_k \, V\!\big(\mathrm{Update}(b,q,y_k)\big)$
        \State $\mathrm{VoI}(q) \gets V_q - V_0 - c(q)$
    \EndFor
    \State $q^{*} \gets \arg\max_{q \in Q} \mathrm{VoI}(q)$
    \If{$\mathrm{VoI}(q^*) \le 0$}
        \textbf{break} \Comment{clarification not worthwhile}
    \Else
        \State Ask $q^*$, observe $y$; $H \gets H \cup \{(q^*,y)\}$; $b \gets \mathrm{Update}(b,q^*,y)$
    \EndIf
\EndFor
\State \textbf{return} $a^* \in \arg\max_{a \in \mathcal{A}} \mathbb{E}_{\theta \sim b}[U(\theta,a)]$ \Comment{final commitment}
\end{algorithmic}
\end{algorithm*}

\section{Experimental Setup}
\subsection{Baseline Methods}
\label{baselines}

\noindent\textbf{No-Question.} This baseline represents the standard agent paradigm. Given the initial query~$S$, the agent commits to an action immediately without any communication with the user. It relies solely on its initial understanding of the user's intent.

\paragraph{Fixed-Round.}
This non-adaptive baseline asks a fixed number of $k$ questions before committing to an action. It serves to isolate the benefit of interaction from the benefit of \textit{adaptive} interaction by exploring a fixed trade-off between information gathering and communication cost.


\paragraph{Adaptive Prompting.}
This baseline prompts the LLM to reason about whether it feels confident enough to act or if it should ask a question. The number of questions is not predetermined, but the decision to stop is based on the model's heuristic self-assessment rather than a formal criterion.


\paragraph{Confidence Thresholding.}
This adaptive baseline formalizes the heuristic of Adaptive Prompting. The agent continues to ask questions as long as its predictive confidence in the best action $a^*$ remains below a tunable threshold $\tau$. We measure confidence using the model's verbalized confidence scores~\citep{tian2023just, zhang2024atomic}, a common practice for modern LLMs. This method is adaptive, but crucially, the threshold $\tau$ must be manually tuned for each task and cost setting to achieve optimal performance.


\subsection{Tasks and Models}
\paragraph{Mixed-Stakes 20 Questions.}

The 20 Questions game is a classic guessing game with a long history as a paradigm for studying human and artificial decision-making under uncertainty. It provides a controlled environment to test how an agent performs strategic information gathering. Following the setup of~\citet{hu2024uncertainty}, the agent must identify a target concept from a known candidate set by asking a series of binary (yes/no) questions. Our key modification is to explicitly test how the agent adapts to varying \textbf{task risk}. We create two parallel versions of this task:

\begin{itemize}
    \item \textbf{Low-Stakes (Animal Guessing):} The agent identifies an animal from a set of 100. A correct guess yields a terminal utility of $U=1$.
    \item \textbf{High-Stakes (Medical Diagnosis):} The agent diagnoses a medical condition from a set of 15 diseases, using real doctor-patient chat histories as input. A correct diagnosis yields a utility of $U=10$.
\end{itemize}



\paragraph{Flight Recommendation} 
We adopt a task designed to model the elicitation of multi-faceted user preferences, a common challenge when aligning agents with diverse user values~\cite{dong-etal-2025-personalization}. Our setup is inspired by the recent work of \cite{qiu2025bayesian} is derived from the FLIGHTPREF dataset originally proposed by \citet{lin2022inferring}. The agent is presented with a user's choice history over five rounds of flight selections. In a final, held-out round, the agent must predict which of three new flight options the user will prefer. Each flight is defined by 8 features (e.g., price, stops, airline), and each user has a latent reward function defining their preferences over these features. The agent can ask clarifying questions to uncover these preferences before making its final prediction. This task tests the agent's ability to strategically query a complex, multi-attribute preference space to infer a user's reward model from their contextual choices. The agent's prediction for the new round will be scored based on this reward function.

\paragraph{Ambiguous WebShop}

To test our agent in a more realistic, interactive environment, we adapt the WebShop benchmark~\citep{yao2022webshop}. In the original setting, user instructions are created to be relatively well-specified (e.g., ``buy a red Adidas t-shirt, size medium''). We deliberately introduce \textbf{query ambiguity} by removing details from the user's request (e.g., ``buy a t-shirt'') to simulate underspecified real-world user query. The agent must then decide whether to act on this partial information (e.g., \verb|search("t-shirt")|) or to ask clarifying questions about attributes like size, color, or brand. This task evaluates the agent's ability to balance autonomous web navigation with strategic information gathering to resolve under-specified user requests. We use GPT-4o to provide a score $\in[0,1]$ for the purchased product against the ground-truth product provided in ~\citet{yao2022webshop}.

\paragraph{Models} We consider a selection of leading LLMs to evaluate the performance of our proposed method, including GPT-4.1 \citep{openai2025gpt41} and Gemini-2.5-Flash \citep{comanici2025gemini}. 


\section{Results}
\subsection{Main Results}

\begin{figure*}[ht!]
    \centering
    \vspace{-10pt}
    \includegraphics[width=0.65\textwidth]{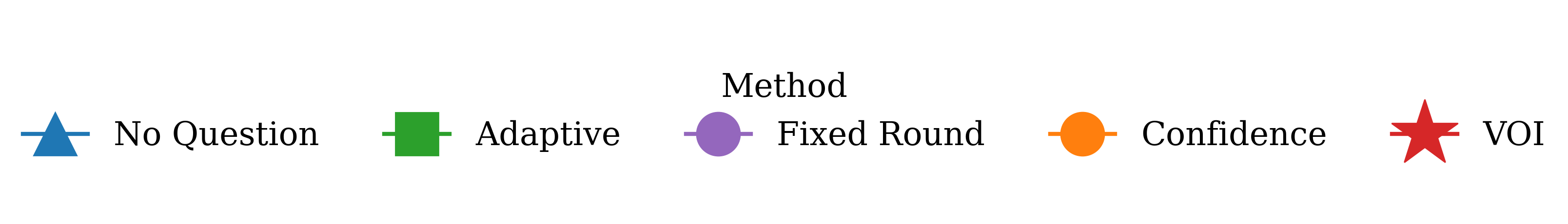}
    \vspace{-8pt}
    
    \subfloat[]{\includegraphics[width=0.32\textwidth]{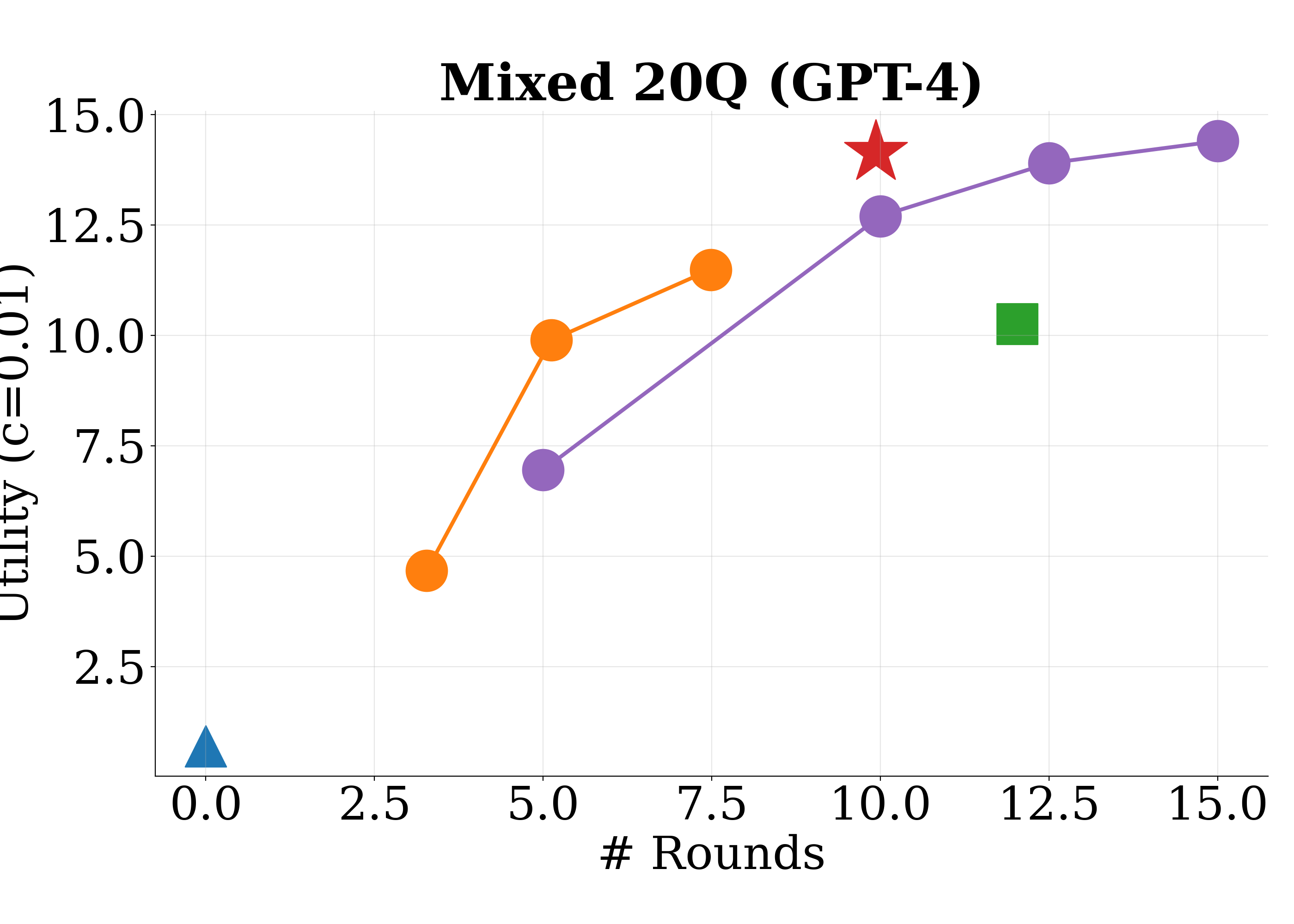}\hspace{4pt}}
    \subfloat[]{\includegraphics[width=0.32\textwidth]{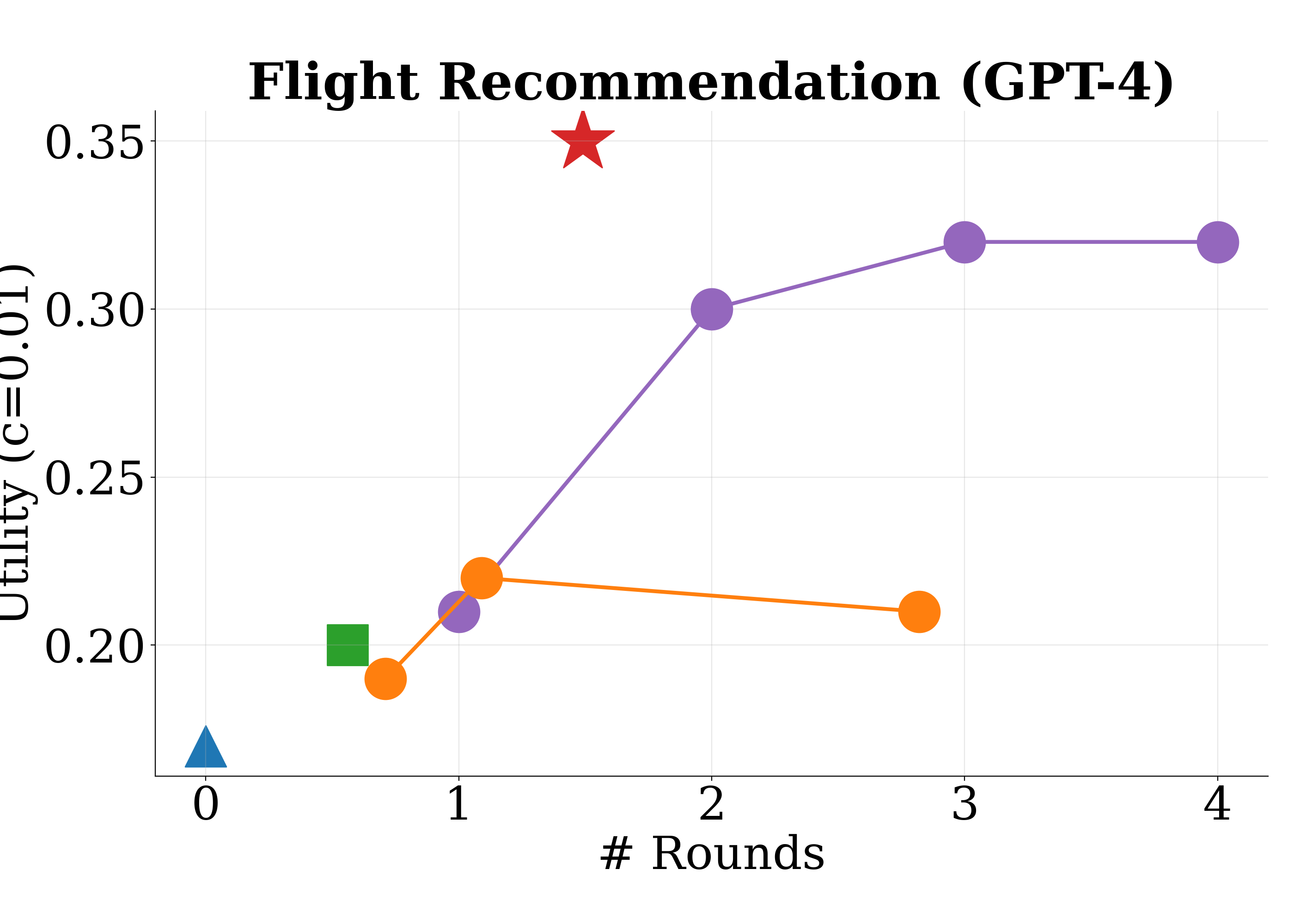}\hspace{4pt}}
    \subfloat[]{\includegraphics[width=0.32\textwidth]{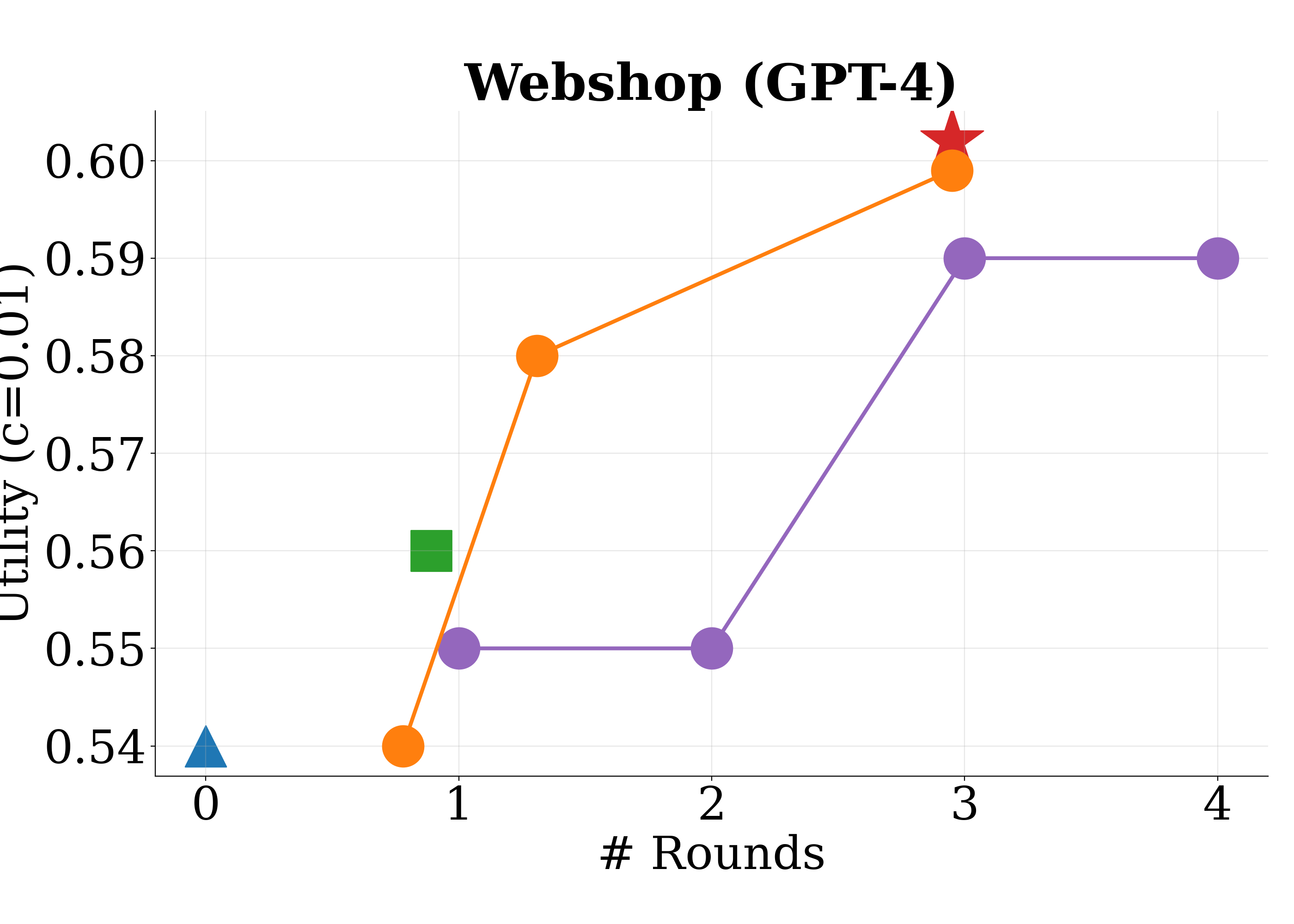}}

    \subfloat[]{\includegraphics[width=0.32\textwidth]{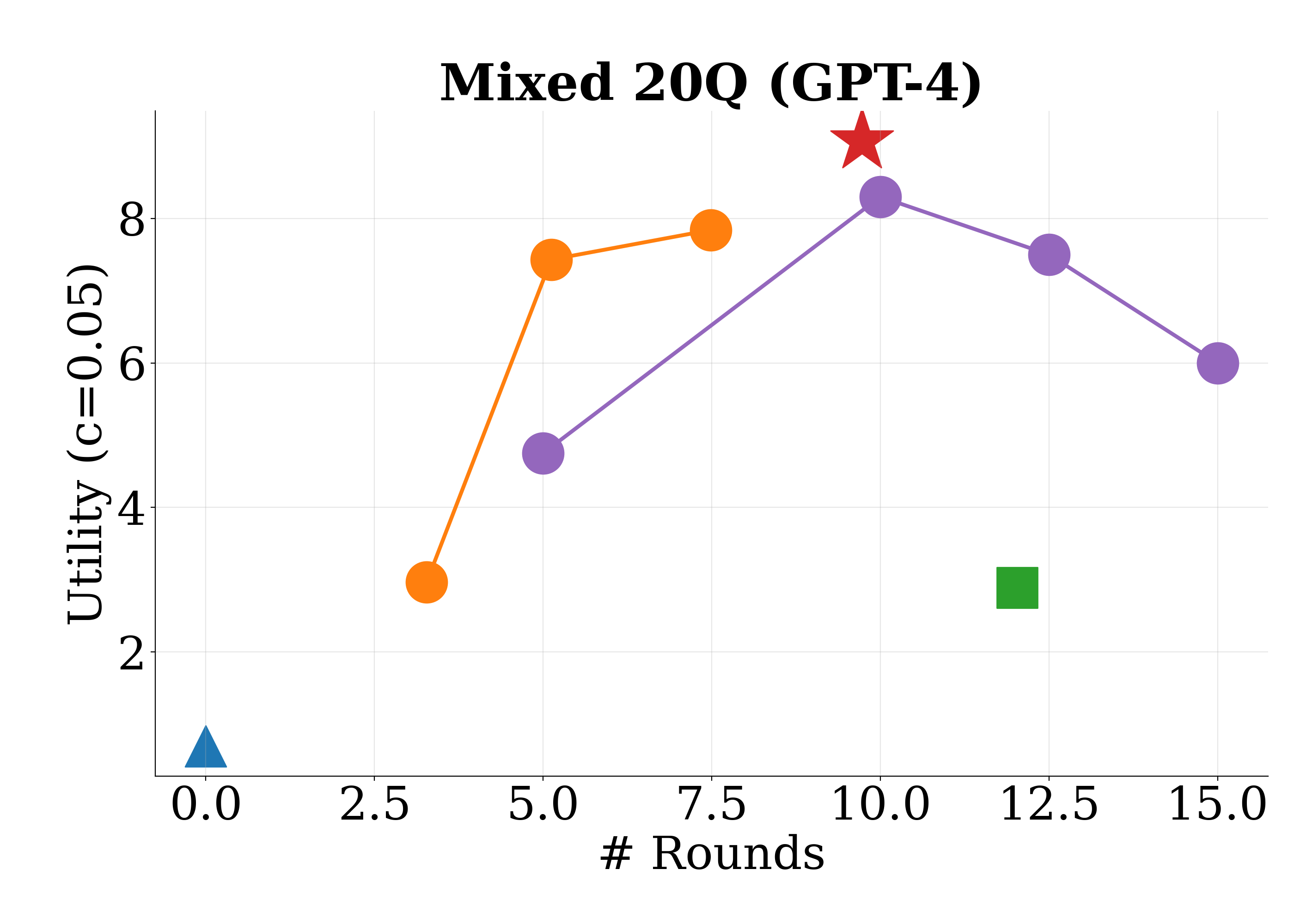}\hspace{4pt}}
    \subfloat[]{\includegraphics[width=0.32\textwidth]{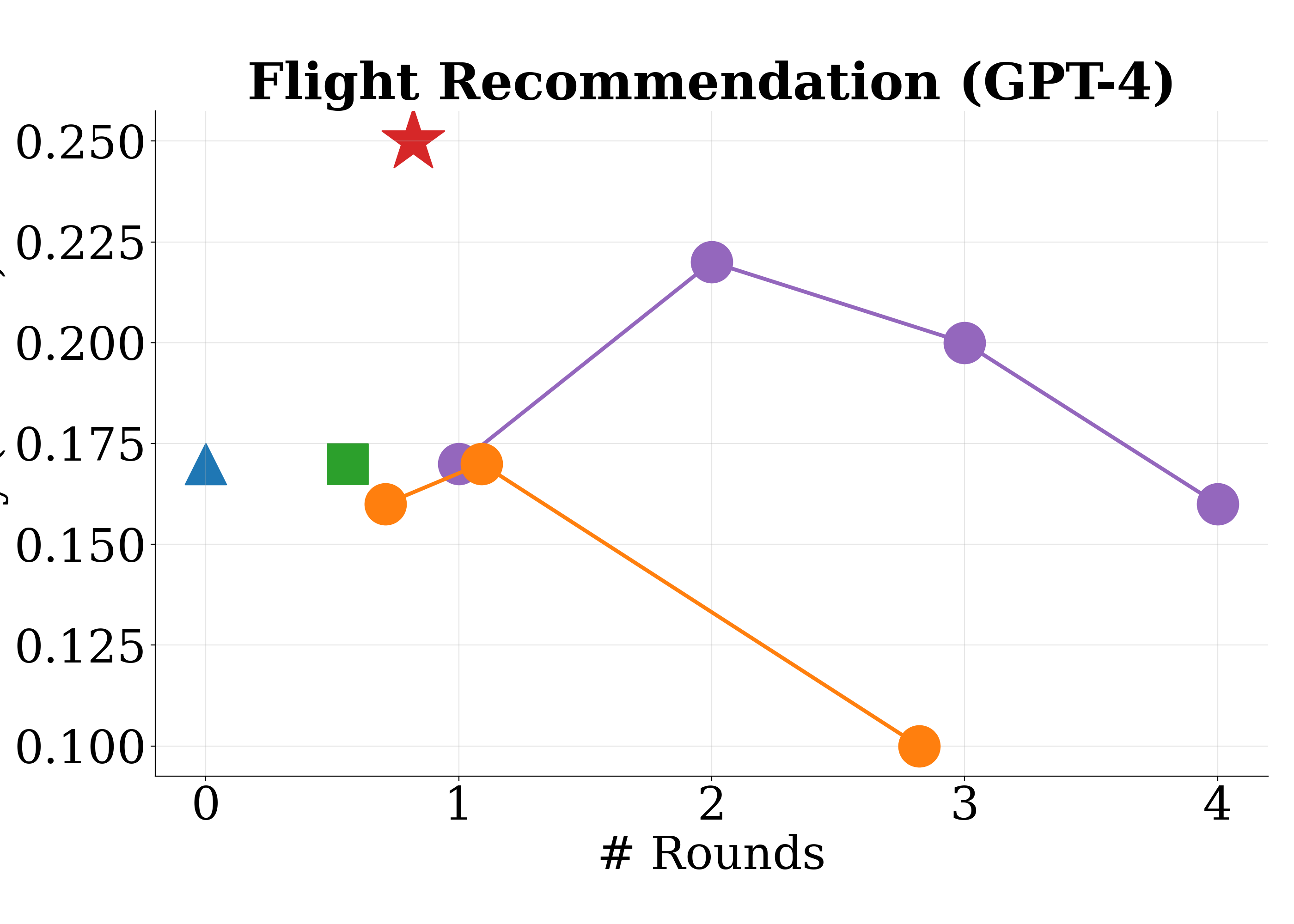}\hspace{4pt}}
    \subfloat[]{\includegraphics[width=0.32\textwidth]{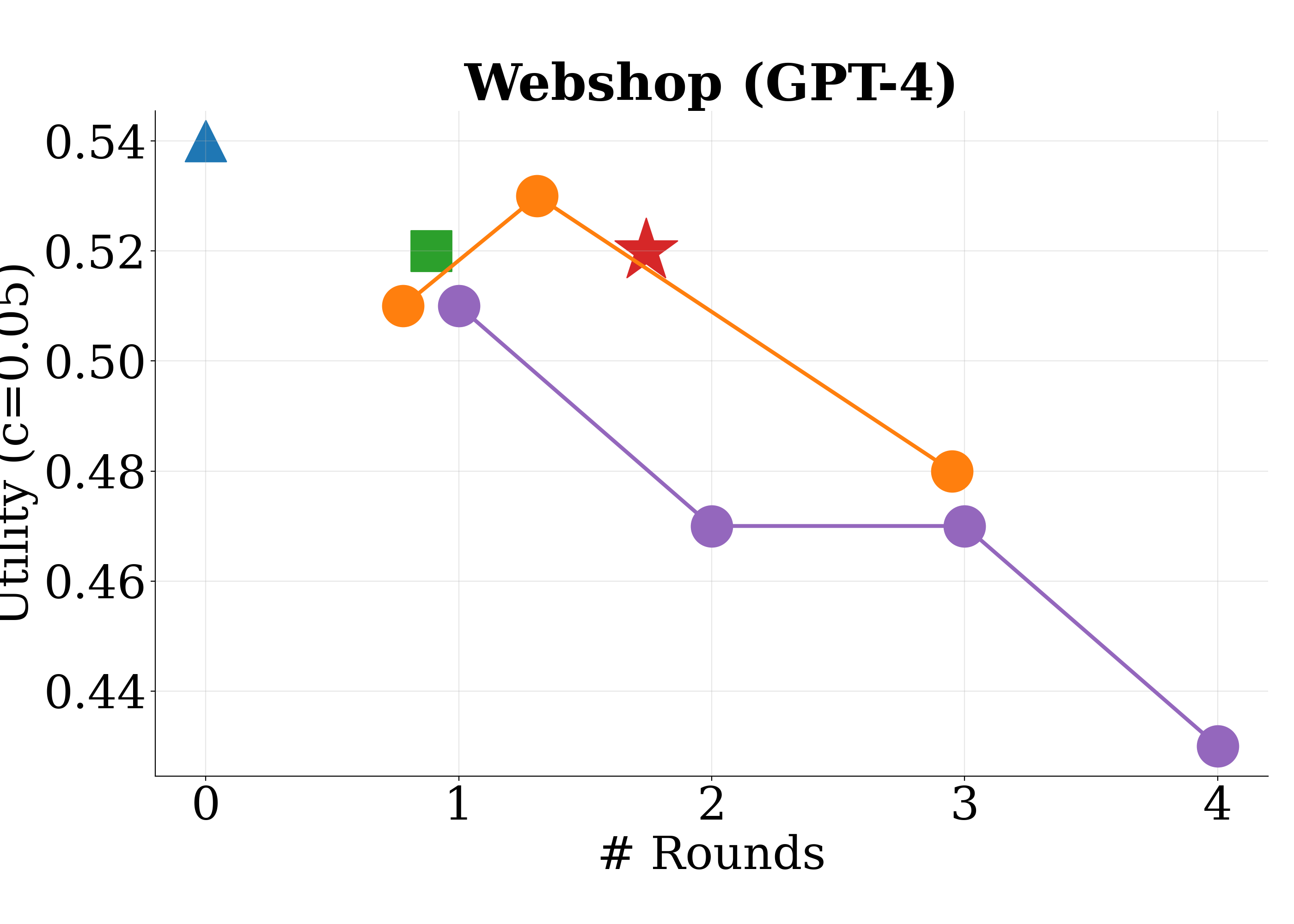}}

    \subfloat[]{\includegraphics[width=0.32\textwidth]{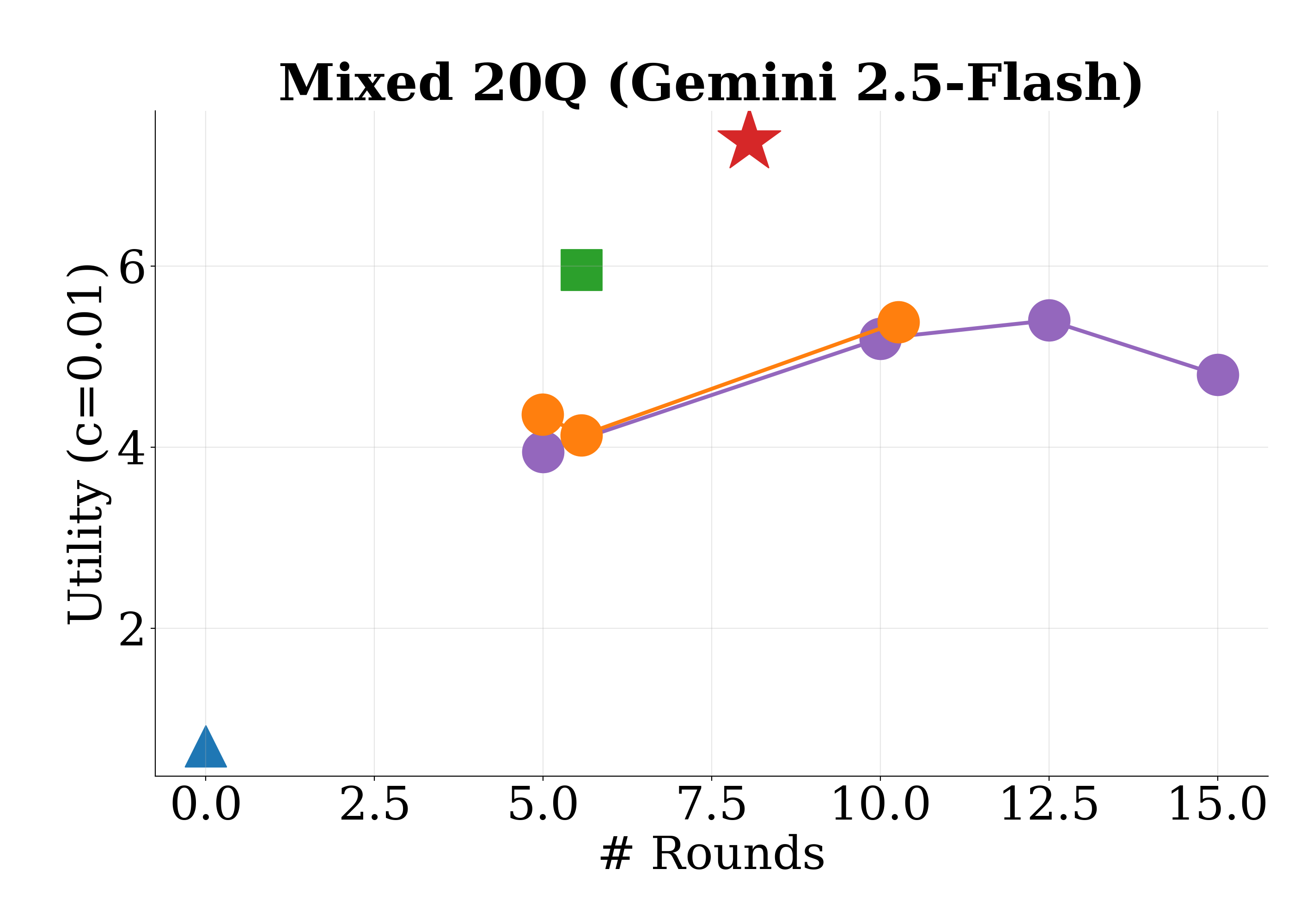}\hspace{4pt}}
    \subfloat[]{\includegraphics[width=0.32\textwidth]{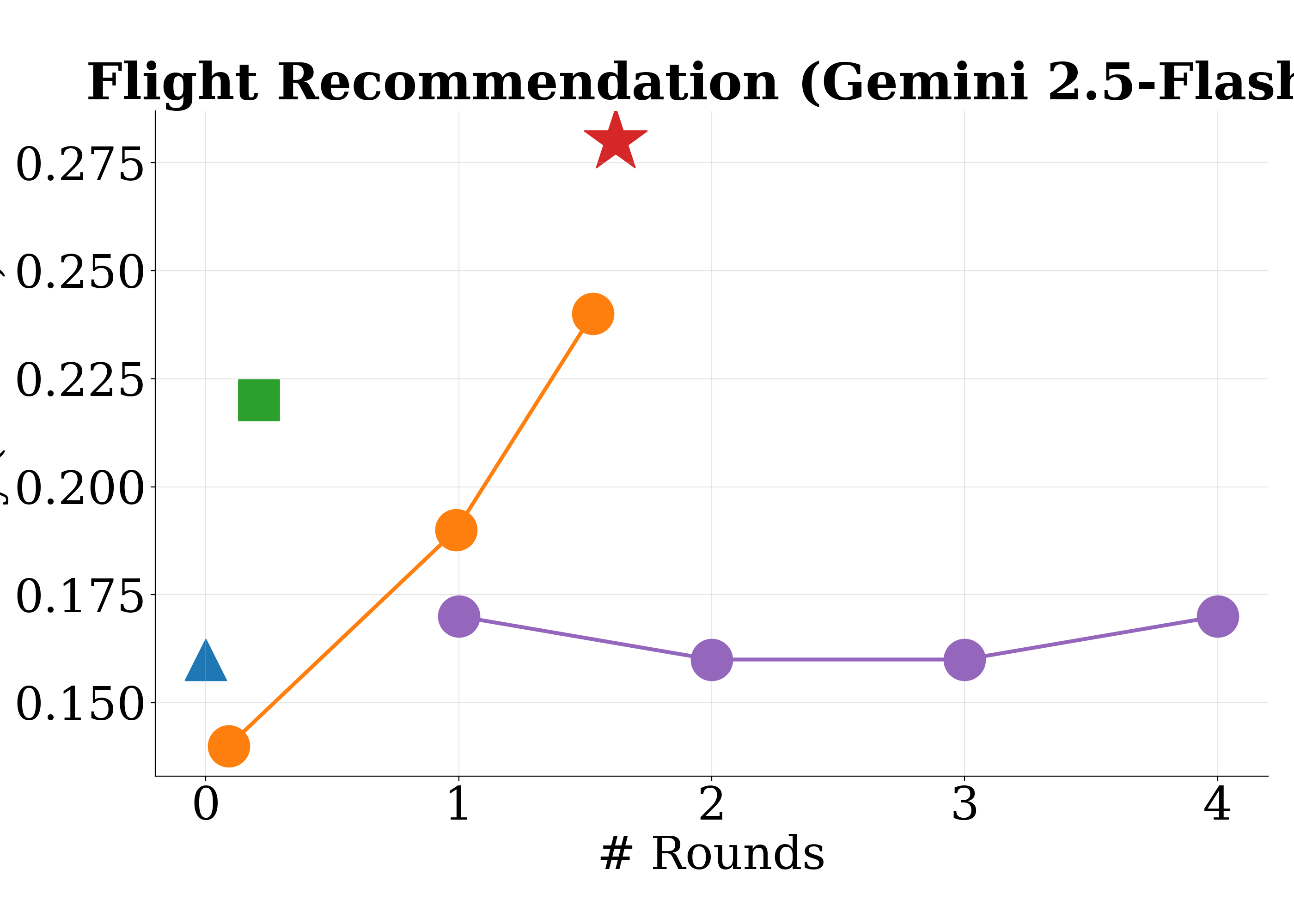}\hspace{4pt}}
    \subfloat[]{\includegraphics[width=0.32\textwidth]{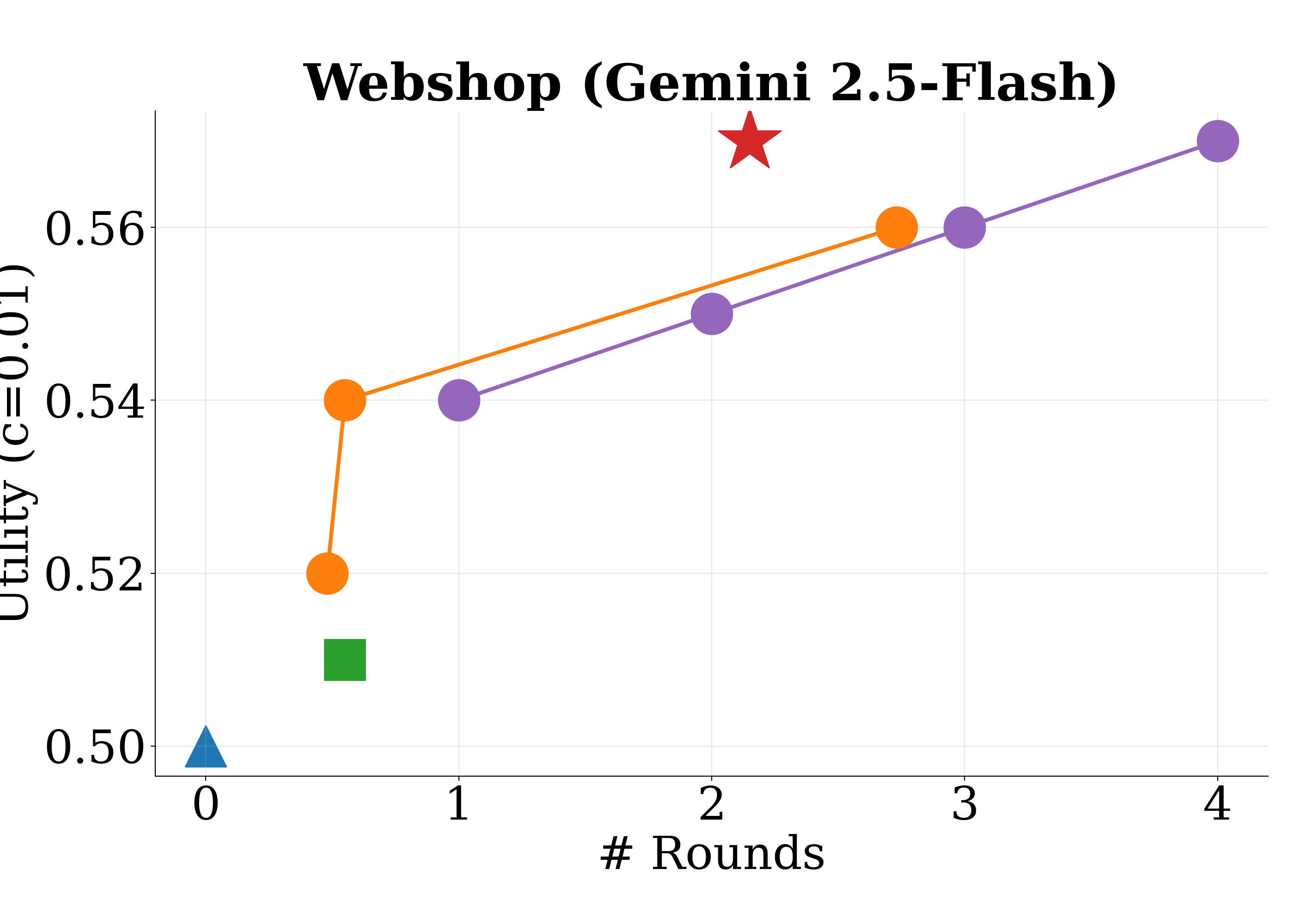}}

    \subfloat[]{\includegraphics[width=0.32\textwidth]{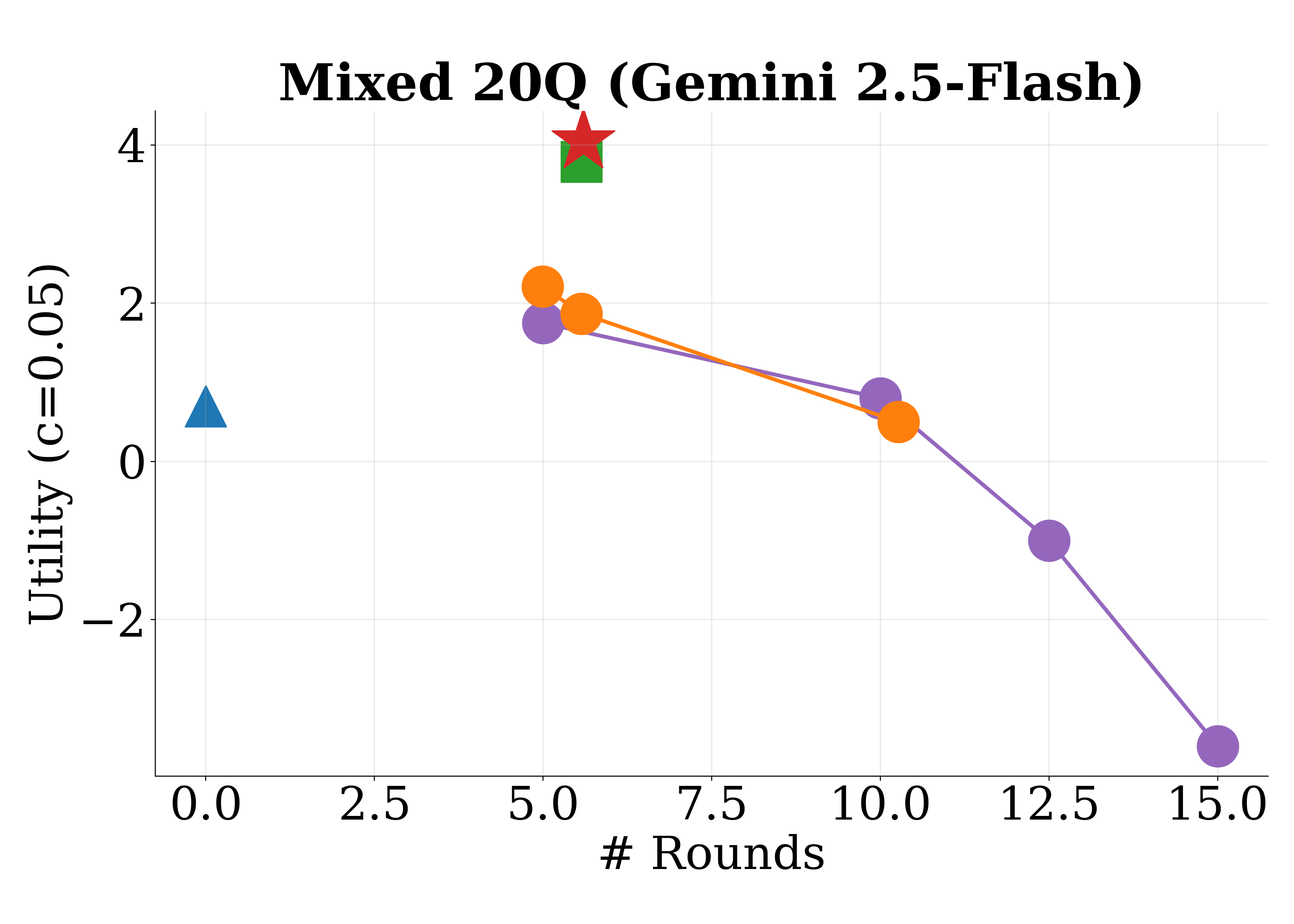}\hspace{4pt}}
    \subfloat[]{\includegraphics[width=0.32\textwidth]{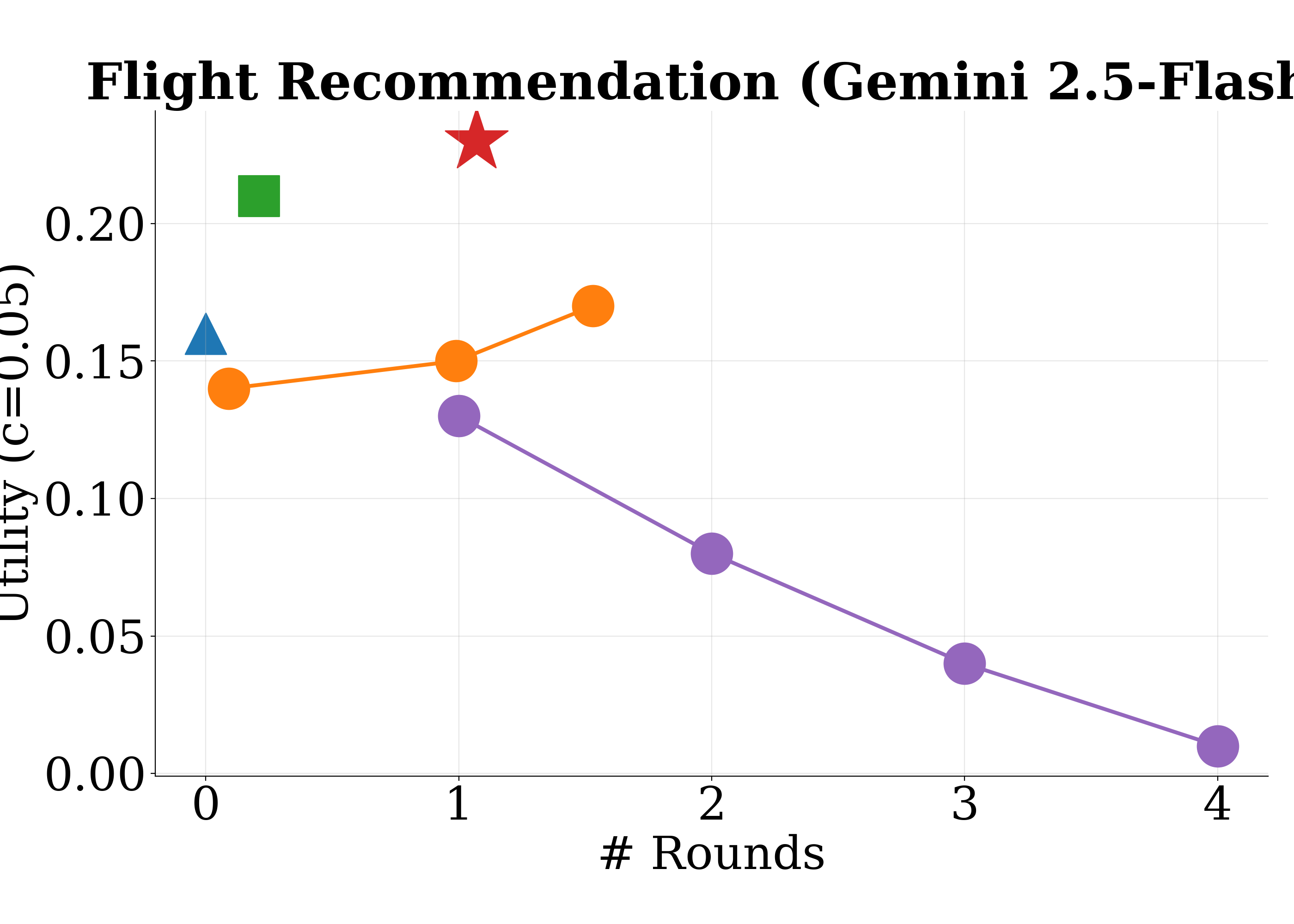}\hspace{4pt}}
    \subfloat[]{\includegraphics[width=0.32\textwidth]{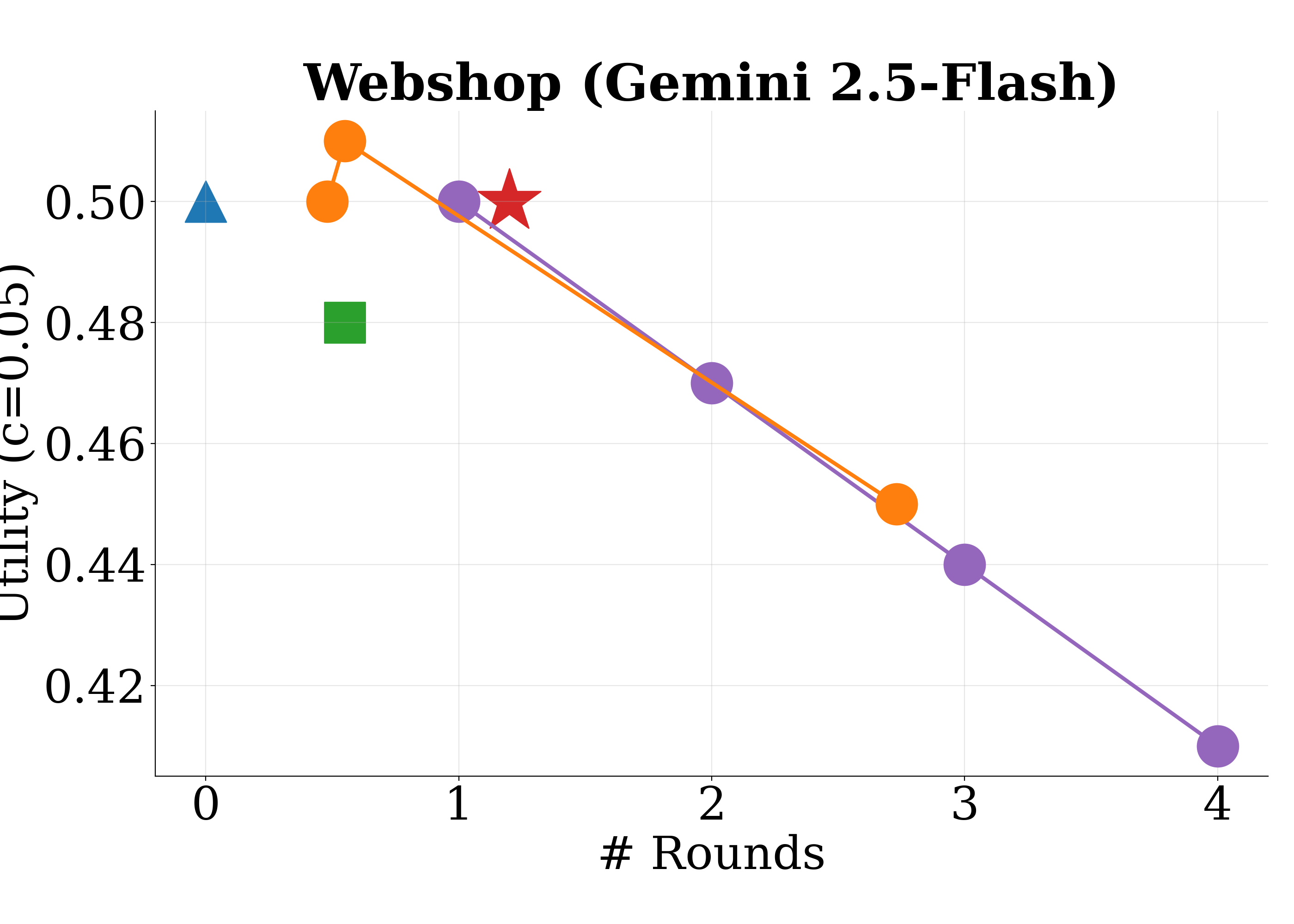}}
    
    \caption{\textbf{Utility vs. Communication Rounds.} Final utility as a function of the number of clarification questions asked across our three tasks, for GPT-4 (top two rows) and Gemini-2.5-Flash (bottom two rows), with communication costs $c=0.01$ and $c=0.05$. Utility is defined as $U(\theta,a) - T \cdot c$. The curves for Fixed Round and Confidence Thresholding represent Pareto frontiers generated by varying their respective hyperparameters ($k$ and $\tau$). In contrast, our VoI agent (starred) is a parameter-free method. In nearly all settings, VoI automatically identifies an operating point that matches or exceeds the performance of the best-tuned baseline, demonstrating its superior adaptability and practical value.}
    \label{fig:main_results}
    \vspace{-10pt}
\end{figure*}

\begin{table*}[ht!]
\small
\centering
\caption{\textbf{VOI vs. Baselines Across Costs (Gemini-2.5-Flash, Mixed 20 Question)}. This table compares the VOI policy's expected reward ($r_{\text{VOI}}$) against the best and second-best baselines via grid searching over 9 values. The $\Delta$ columns report VOI's margin over each baseline (positive means VOI is better).}
\begin{tabular}{c | c c | c c | c | c c}
\hline
Cost & Best Baseline & $r_{\text{max}}$ & Second Best & $r_{\text{second}}$ & $r_{\text{VOI}}$ & $r_{\text{VOI}}$--$r_{\text{max}}$  & $r_{\text{VOI}}$--$r_{\text{second}}$\\
\hline
0.01 & Confidence ($\tau$=0.9) & 8.30 & Round ($\tau$=15) & 8.10 & 8.64 & 0.34 & 0.54 \\
0.02 & Confidence ($\tau$=0.9) & 6.88 & Confidence ($\tau$=0.9) & 6.80 & 7.72 & 0.84 & 0.92 \\
0.05 & Round ($\tau$=5)  & 3.65 & Confidence ($\tau$=0.5) & 3.64 & 5.01 & 1.36 & 1.37 \\
0.10 & Confidence ($\tau$=0.5) & 2.28 & Round ($\tau$=5)  & 0.90 & 1.38 & -0.90 & 0.48 \\
0.20 & No Question & 0 & Round ($\tau$=5) & -4.60 & -0.96 & -0.96 & 3.64 \\
\hline
\end{tabular}
\label{tab: ablation cost}
\end{table*}

\begin{figure*}[ht!]
    \begin{subfigure}{0.33\linewidth}
        \includegraphics[width=\linewidth]{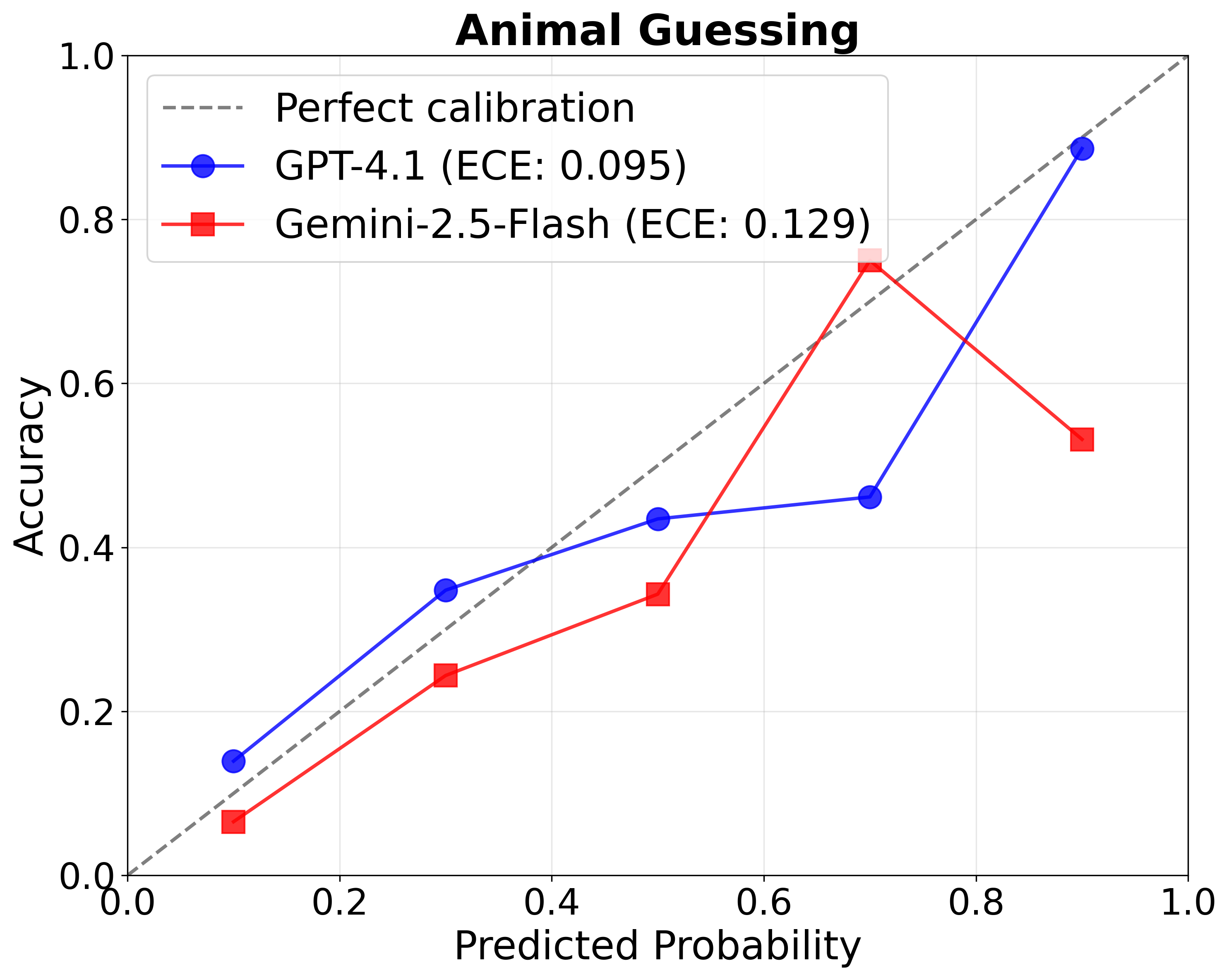}
        \caption{}
    \end{subfigure}\hfill
    \begin{subfigure}{0.33\linewidth}
        \includegraphics[width=\linewidth]{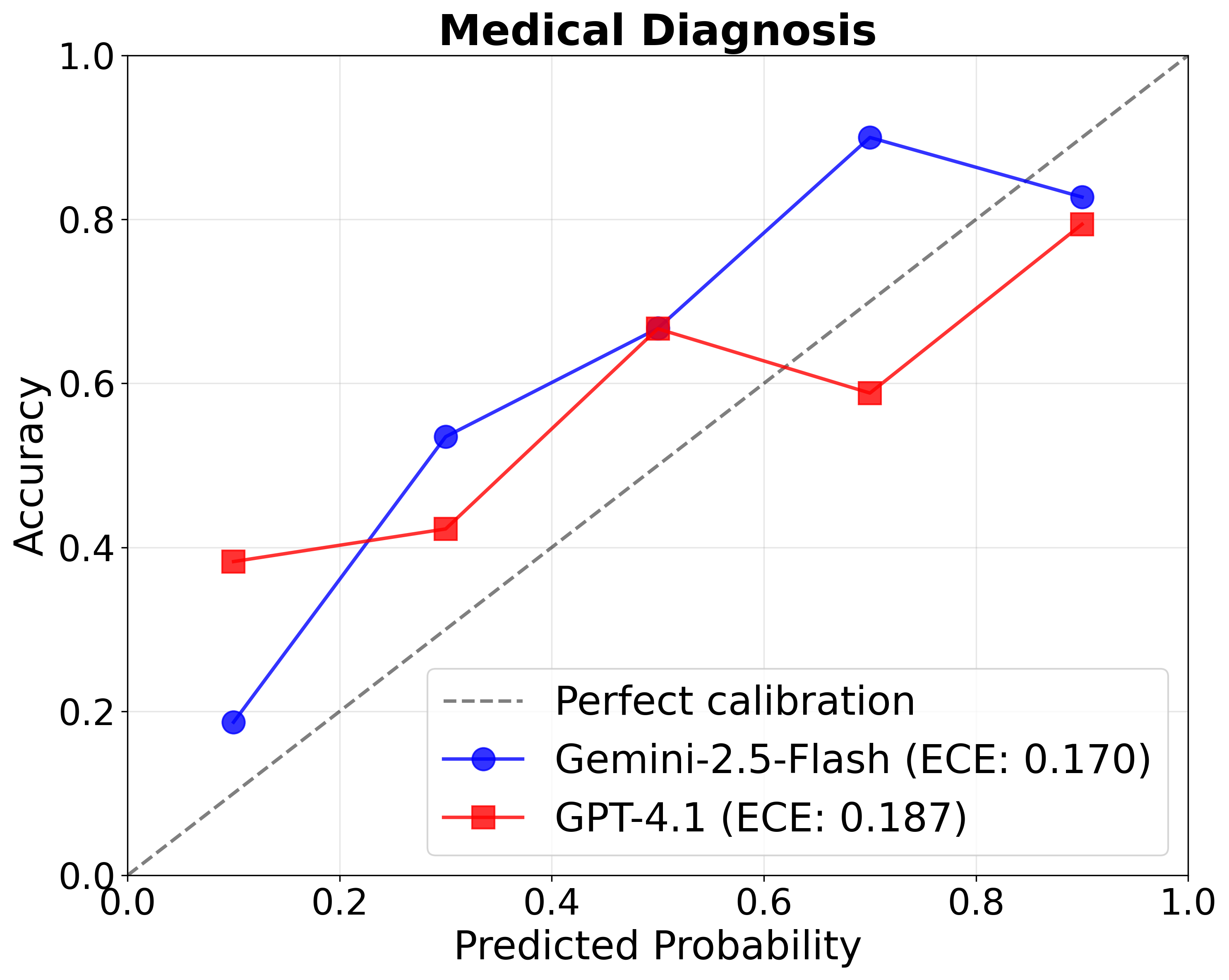}
        \caption{}
    \end{subfigure}\hfill
    \begin{subfigure}{0.33\linewidth}
        \includegraphics[width=\linewidth]{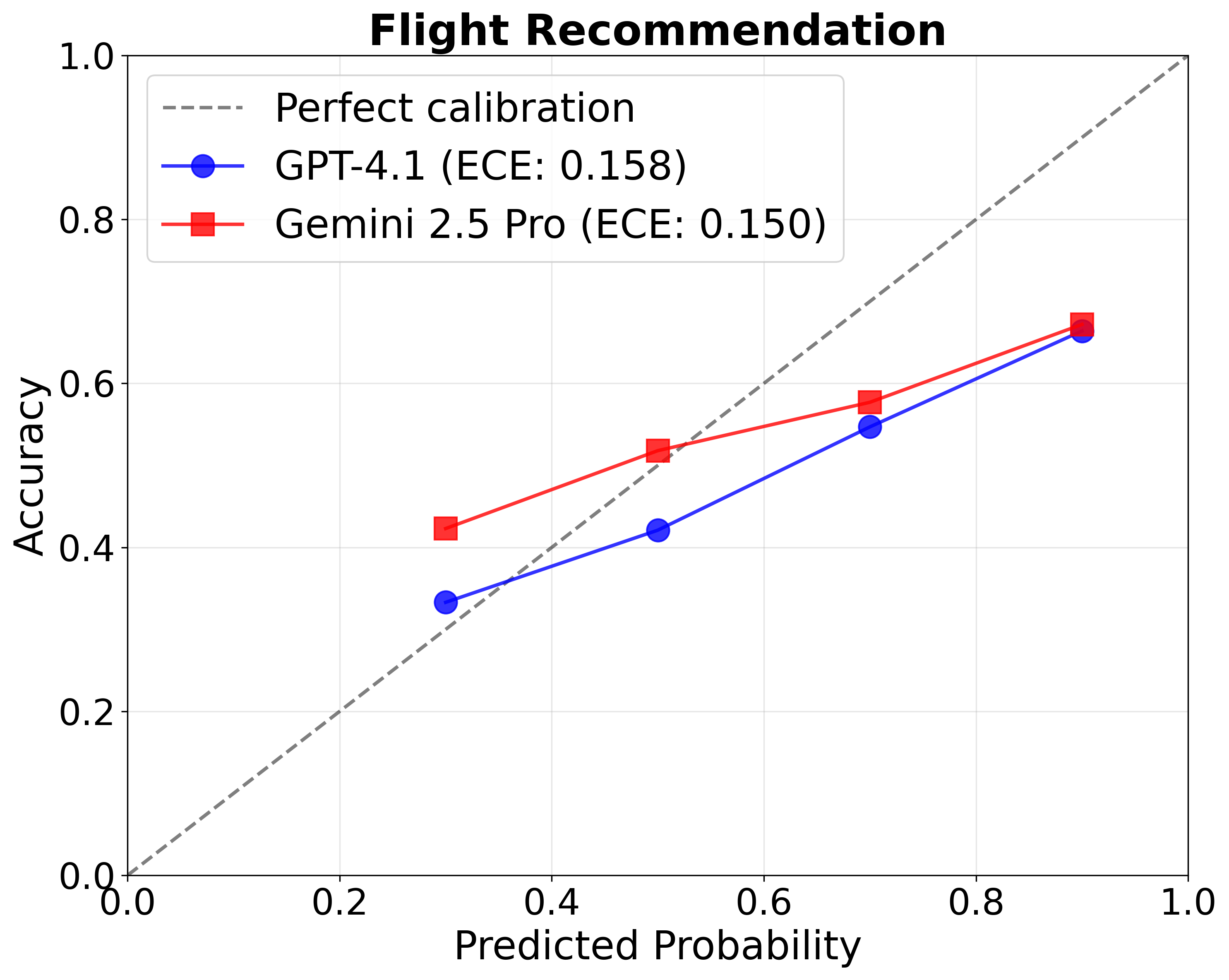}
        \caption{}
    \end{subfigure}
    \caption{\textbf{Calibration Analysis} The figure presents the calibration analysis of GPT-4 and Gemini-2.5-Flash on Animal Guessing, Medical Diagnoiss, and Flight Recommendation. (In (c) the accuracy for predicted probability between 0 and 0.2 is omitted because very few samples fall in that range.}
    \label{fig: calibration}
\end{figure*}



Our central findings are summarized in Figure~\ref{fig:main_results}. Across all tasks and communication cost settings, our VoI-based agent consistently achieves state-of-the-art utility. Crucially, it does so without requiring task-specific threshold tuning, showcasing its robustness and practical advantages.

\paragraph{VoI excels by finding the optimal utility-cost balance.}
As shown in Figure~\ref{fig:main_results}, our VoI agent (starred marker) consistently ranks as the top-performing method across the Mixed 20Q, Flight Recommendation, and Ambiguous WebShop tasks. For instance, in Mixed 20Q with a communication cost of $c=0.01$, VOI achieves a utility of 14.14, significantly outperforming the best-tuned confidence-thresholding baseline (11.49 at $\tau=0.90$). This performance advantage stems from VOI's ability to dynamically determine the optimal number of clarification questions, a stark contrast to fixed-round and confidence-based methods that require brittle, manual tuning of a threshold for each specific task and cost structure.

\paragraph{Adaptive communication is essential for ambiguous tasks.}
The ``No Question'' baseline establishes the necessity of proactive communication. On the Mixed 20Q task, where the initial query is inherently underspecified, this baseline's accuracy is near zero for both low-stakes (animal) and high-stakes (medical) variants. However, as shown in Figures~\ref{fig:main_results}(f) and~\ref{fig:main_results}(l), when communication costs are prohibitively high, avoiding questions becomes a competitive strategy. In these scenarios, our VOI method correctly adapts by stopping communication early, demonstrating its ability to gracefully handle the full spectrum of cost-benefit scenarios.

%
\paragraph{Adaptive prompting are insufficient for robust performance.}
The Adaptive Prompting baseline shows that simply instructing an LLM to ``ask questions when needed'' offers an improvement over non-adaptive strategies. However, its performance is inconsistent and consistently lower than more structured methods. This is because the decision to communicate is based on the model’s internal “feeling” of confidence, which is often poorly calibrated~\cite{hu2025navigating,zhang2026confidence}, rather than a formal criterion. It lacks a principled mechanism to weigh the potential information gain against the explicit communication cost, leading to suboptimal and unpredictable behavior.

\paragraph{Fixed-round communication strategies are fundamentally suboptimal.}
A fixed-round policy, which asks a predetermined number of questions, fails to adapt to the specific needs of a given query. As illustrated in the inverted-\emph{U} shape of the ``Fixed Round'' curves in Figure~\ref{fig:main_results}, utility initially increases with more questions but then declines as communication costs overwhelm the benefits of additional information. The optimal number of questions varies significantly with the task and cost, highlighting the necessity of an adaptive policy.

\paragraph{Confidence thresholding is effective but brittle.}
The confidence thresholding baseline provides a strong, adaptive competitor. With the \emph{correctly} tuned confidence threshold $\tau$, its performance can be comparable to our VOI method (e.g., on GPT-4 for Mixed 20Q and Webshop). However, this effectiveness is its Achilles' heel; the optimal $\tau$ is highly sensitive and must be manually selected for each task and cost combination, making it impractical for real-world deployment. Our VoI method provides a principled solution that matches or exceeds this performance without any such manual tuning.

\subsection{Ablation Study}
\paragraph{Ablation on Communication Cost.}
As shown in Table~\ref{tab: ablation cost}, across the cost sweep on Mixed 20-Question the VoI controller matches or exceeds the strongest grid-searched baselines. We tune four baselines over nine threshold settings, and while the best baseline shifts with the communication cost, VoI consistently selects an appropriate number of questions thst match the performance of the best baseline. Importantly, this pattern is stable across different choice of communication costs: VoI adapts smoothly to the stated cost rather than hinging on a brittle threshold choice.

\paragraph{Calibration Analysis.}  A critical component of our VoI framework is the LLM's ability to estimate a belief distribution $b(\theta)$ over latent user states. To analyze it, ideally we should compare model predicted distribution to the ground truth distribution. However, in the absence of the ground truth distribution for our tasks, we instead measure the argmax from the distribution against the ground truth item as the standard calibration analysis to approximate its distribution estimation accuracy. As shown in Figure \ref{fig: calibration}, The results reveal that models are reasonably calibrated in Animal Guessing game but less calibrated for Medical Diagnosis which we suspect because of the inherent complication and noise in the symptoms of diseases. Despite this, we see that VOI are empirically effective and robust that consistently matches if not perform the best baselines after searching hyperparameters. We believe that current and future work that could improving model calibration under missing context \cite{li2025semantic,zhang2026confidence} could further improve the performance of VOI.



\section{Conclusion} 


Current LLM agents are often designed for well-specified tasks, leaving them brittle when faced with the inherent ambiguity of real-world user requests. In this work, we argued that overcoming this limitation requires agents to move beyond simple execution and develop a principled strategy for adaptive communication. We proposed a formal framework for this problem, centered on balancing three key factors: query ambiguity, task risk, and user cognitive load.
Our primary contribution is a practical, inference-time method based on the Value of Information (VoI) that operationalizes this framework. By explicitly calculating the expected utility gain of a potential question and weighing it against its communication cost, our VoI-driven agent decides when to act and when to ask. Extensive experiments across diverse domains—including medical diagnosis and online shopping—demonstrate that our approach consistently outperforms non-adaptive and heuristic-based baselines. Crucially, it achieves this without the need for the brittle, task-specific threshold tuning that plagues other adaptive methods.
Ultimately, this work provides a principled foundation for building LLM agents that are not just capable executors, but also thoughtful communicators. By equipping agents with a formal understanding of when information is valuable, we can create more aligned, efficient, and truly collaborative human-AI systems.

\section*{Limitations}

\paragraph{Scope of Interaction: Decision vs. Generation.}
Our work focuses on the core decision of \textit{when} to communicate, rather than \textit{what} questions to generate. To this end, our experiments utilize a predefined set of actions ($a \in \mathcal{A}$) and clarifying questions, a methodological choice consistent with prior work~\citep{hu2024uncertainty, kobalczyk2025active}. This controlled setting isolates the performance of our VoI-based \textit{selection policy}, providing an unambiguous evaluation of our central claim. By controlling for the quality of question generation, we demonstrate the effectiveness of the decision-making principle itself. Extending this framework to fully open-ended dialogue is an important next step; establishing this selection principle is a necessary foundation. Our work provides the core engine around which more sophisticated generative components can be built.

\paragraph{Model of Communication Cost.}
We employ a linear communication cost model ($c(H) = T \cdot c$). Accurately modeling the nuances of human cognitive load is a major, open research challenge in its own right, spanning HCI and cognitive science. Therefore, in line with common practice in decision-theoretic analyses, we adopt a simplified and interpretable cost function. This allows us to clearly illustrate the fundamental trade-off between utility gain and cost, without introducing confounding variables from a more complex, speculative cognitive model. Importantly, the VoI framework itself is agnostic to the form of the cost function; the core decision rule, $\text{VoI}(q) - c(H)$, can readily incorporate more sophisticated models as they are developed. We view the linear cost model as a reasonable first-order approximation that demonstrates the framework's viability, with refinement through empirical user research as a natural next step.

\section*{Ethical Considerations}
While our VoI framework optimizes the trade-off between information gain and communication cost, user agency must remain paramount: users should retain the ability to decline questions or proceed without clarification based on their own judgment. Beyond this, the act of questioning introduces critical considerations regarding user burden and privacy. First, clarifying questions—even when theoretically optimal—inherently impose a cognitive demand on the user; an agent that queries too frequently or intrusively risks eroding trust and causing frustration, necessitating cost models that strictly penalize unnecessary interruptions. Second, the pursuit of resolving ambiguity often requires eliciting specific, potentially sensitive information (e.g., medical symptoms or personal preferences) to update the agent's belief distribution. It is imperative that future implementations incorporate strict data minimization principles and privacy safeguards to ensure that the agent's drive for reduced uncertainty does not compromise user privacy or comfort \cite{hui2025toxic,hui2025privacypad, dong2025undial}. We acknowledge the use of AI tools for refining the paper writing.

\section*{Acknowledgements}
T.H is supported by Gates Cambridge Trust (grant OPP1144 from the Bill \& Melinda Gates Foundation). This work was partially performed using resources provided by the Cambridge Service for Data Driven Discovery (CSD3) operated by the University of Cambridge Research Computing Service (www.csd3.cam.ac.uk), provided by Dell EMC and Intel using Tier-2 funding from the Engineering and Physical Sciences Research Council (capital grant EP/T022159/1), and DiRAC funding from the Science and Technology Facilities Council (www.dirac.ac.uk).



\bibliography{main}

@article{zhang2023clarify,
 author = {Zhang, Michael JQ and Choi, Eunsol},
 journal = {ArXiv preprint},
 title = {Clarify when necessary: Resolving ambiguity through interaction with lms},
 url = {https://arxiv.org/abs/2311.09469},
 volume = {abs/2311.09469},
 year = {2023}
}

@article{zhang2024atomic,
  title={Atomic calibration of llms in long-form generations},
  author={Zhang, Caiqi and Yang, Ruihan and Zhang, Zhisong and Huang, Xinting and Yang, Sen and Yu, Dong and Collier, Nigel},
  journal={arXiv preprint arXiv:2410.13246},
  year={2024}
}

@article{hui2025privacypad,
  title={PrivacyPAD: A Reinforcement Learning Framework for Dynamic Privacy-Aware Delegation},
  author={Hui, Zheng and Dong, Yijiang River and Sivapiromrat, Sanhanat and Shareghi, Ehsan and Collier, Nigel},
  journal={arXiv preprint arXiv:2510.16054},
  year={2025}
}

@inproceedings{dong2025undial,
  title={Undial: Self-distillation with adjusted logits for robust unlearning in large language models},
  author={Dong, Yijiang River and Lin, Hongzhou and Belkin, Mikhail and Huerta, Ramon and Vuli{\'c}, Ivan},
  booktitle={Proceedings of the 2025 Conference of the Nations of the Americas Chapter of the Association for Computational Linguistics: Human Language Technologies (Volume 1: Long Papers)},
  pages={8827--8840},
  year={2025}
}

@misc{openai2025gpt41,
  author       = {OpenAI},
  title        = {Introducing GPT-4.1 in the API},
  year         = {2025},
  month        = {April},
  day          = {14},
  url          = {https://openai.com/index/gpt-4-1/},
  note         = {Accessed: 2025-09-18}
}

@article{comanici2025gemini,
  title={Gemini 2.5: Pushing the frontier with advanced reasoning, multimodality, long context, and next generation agentic capabilities},
  author={Comanici, Gheorghe and Bieber, Eric and Schaekermann, Mike and Pasupat, Ice and Sachdeva, Noveen and Dhillon, Inderjit and Blistein, Marcel and Ram, Ori and Zhang, Dan and Rosen, Evan and others},
  journal={arXiv preprint arXiv:2507.06261},
  year={2025}
}

@article{lin-etal-2024-decision,
 address = {Cambridge, MA},
 author = {Lin, Jessy  and
Tomlin, Nicholas  and
Andreas, Jacob  and
Eisner, Jason},
 doi = {10.1162/tacl_a_00679},
 journal = {Transactions of the Association for Computational Linguistics},
 pages = {892--911},
 publisher = {MIT Press},
 title = {Decision-Oriented Dialogue for Human-{AI} Collaboration},
 url = {https://aclanthology.org/2024.tacl-1.50/},
 volume = {12},
 year = {2024}
}

@inproceedings{hui-etal-2025-winspot,
    title = "{W}in{S}pot: {GUI} Grounding Benchmark with Multimodal Large Language Models",
    author = "Hui, Zheng  and
      Li, Yinheng  and
      Zhao, Dan  and
      Banbury, Colby  and
      Chen, Tianyi  and
      Koishida, Kazuhito",
    editor = "Che, Wanxiang  and
      Nabende, Joyce  and
      Shutova, Ekaterina  and
      Pilehvar, Mohammad Taher",
    booktitle = "Proceedings of the 63rd Annual Meeting of the Association for Computational Linguistics (Volume 2: Short Papers)",
    month = jul,
    year = "2025",
    address = "Vienna, Austria",
    publisher = "Association for Computational Linguistics",
    url = "https://aclanthology.org/2025.acl-short.85/",
    doi = "10.18653/v1/2025.acl-short.85",
    pages = "1086--1096",
    ISBN = "979-8-89176-252-7"
}

@misc{hui2025safe,
      title={Toward Safe and Human-Aligned Game Conversational Recommendation via Multi-Agent Decomposition}, 
      author={Zheng Hui and Xiaokai Wei and Yexi Jiang and Kevin Gao and Chen Wang and Frank Ong and Se-eun Yoon and Rachit Pareek and Michelle Gong},
      year={2025},
      eprint={2504.20094},
      archivePrefix={arXiv},
      primaryClass={cs.IR},
      url={https://arxiv.org/abs/2504.20094}, 
}

@misc{hui2025toxic,
      title={ToxiCraft: A Novel Framework for Synthetic Generation of Harmful Information}, 
      author={Zheng Hui and Zhaoxiao Guo and Hang Zhao and Juanyong Duan and Congrui Huang},
      year={2025},
      eprint={2409.14740},
      archivePrefix={arXiv},
      primaryClass={cs.CL},
      url={https://arxiv.org/abs/2409.14740}, 
}

@article{zhang2024ask,
 author = {Zhang, Xuan and Deng, Yang and Ren, Zifeng and Ng, See-Kiong and Chua, Tat-Seng},
 journal = {ArXiv preprint},
 title = {Ask-before-plan: Proactive language agents for real-world planning},
 url = {https://arxiv.org/abs/2406.12639},
 volume = {abs/2406.12639},
 year = {2024}
}

@article{kobalczyk2025active,
 author = {Kobalczyk, Katarzyna and Astorga, Nicolas and Liu, Tennison and van der Schaar, Mihaela},
 journal = {ArXiv preprint},
 title = {Active Task Disambiguation with LLMs},
 url = {https://arxiv.org/abs/2502.04485},
 volume = {abs/2502.04485},
 year = {2025}
}

@inproceedings{deng2023prompting,
 address = {Singapore},
 author = {Deng, Yang  and
Liao, Lizi  and
Chen, Liang  and
Wang, Hongru  and
Lei, Wenqiang  and
Chua, Tat-Seng},
 booktitle = {Findings of the Association for Computational Linguistics: EMNLP 2023},
 doi = {10.18653/v1/2023.findings-emnlp.711},
 editor = {Bouamor, Houda  and
Pino, Juan  and
Bali, Kalika},
 pages = {10602--10621},
 publisher = {Association for Computational Linguistics},
 title = {Prompting and Evaluating Large Language Models for Proactive Dialogues: Clarification, Target-guided, and Non-collaboration},
 url = {https://aclanthology.org/2023.findings-emnlp.711},
 year = {2023}
}

@article{wang2025adaptive,
 author = {Wang, Jimmy and Zollo, Thomas and Zemel, Richard and Namkoong, Hongseok},
 journal = {ArXiv preprint},
 title = {Adaptive Elicitation of Latent Information Using Natural Language},
 url = {https://arxiv.org/abs/2504.04204},
 volume = {abs/2504.04204},
 year = {2025}
}

@article{li2025semantic,
 author = {Li, Xiaomin and Yu, Zhou and Zhang, Ziji and Zhuang, Yingying and Shah, Swair and Sadagopan, Narayanan and Beniwal, Anurag},
 journal = {ArXiv preprint},
 title = {Semantic volume: Quantifying and detecting both external and internal uncertainty in llms},
 url = {https://arxiv.org/abs/2502.21239},
 volume = {abs/2502.21239},
 year = {2025}
}

@article{shah2024agents,
 author = {Shah, Chirag and White, Ryen W},
 journal = {ArXiv preprint},
 title = {Agents are not enough},
 url = {https://arxiv.org/abs/2412.16241},
 volume = {abs/2412.16241},
 year = {2024}
}

@article{liu2024dellma,
 author = {Liu, Ollie and Fu, Deqing and Yogatama, Dani and Neiswanger, Willie},
 journal = {ArXiv preprint},
 title = {Dellma: Decision making under uncertainty with large language models},
 url = {https://arxiv.org/abs/2402.02392},
 volume = {abs/2402.02392},
 year = {2024}
}

@inproceedings{yao2022webshop,
 author = {Shunyu Yao and
Howard Chen and
John Yang and
Karthik Narasimhan},
 bibsource = {dblp computer science bibliography, https://dblp.org},
 booktitle = {Advances in Neural Information Processing Systems 35: Annual Conference
on Neural Information Processing Systems 2022, NeurIPS 2022, New Orleans,
LA, USA, November 28 - December 9, 2022},
 editor = {Sanmi Koyejo and
S. Mohamed and
A. Agarwal and
Danielle Belgrave and
K. Cho and
A. Oh},
 title = {WebShop: Towards Scalable Real-World Web Interaction with Grounded
Language Agents},
 url = {http://papers.nips.cc/paper\_files/paper/2022/hash/82ad13ec01f9fe44c01cb91814fd7b8c-Abstract-Conference.html},
 year = {2022}
}

@misc{chen_learning_2024,
 author = {Chen, Maximillian and Sun, Ruoxi and Arık, Sercan Ö and Pfister, Tomas},
 journal = {ArXiv preprint},
 title = {Learning to {Clarify}: {Multi}-turn {Conversations} with {Action}-{Based} {Contrastive} {Self}-{Training}},
 url = {https://arxiv.org/abs/2406.00222},
 volume = {abs/2406.00222},
 year = {2024}
}

@misc{zhang_modeling_2024,
 author = {Zhang, Michael J. Q. and Knox, W. Bradley and Choi, Eunsol},
 journal = {ArXiv preprint},
 title = {Modeling {Future} {Conversation} {Turns} to {Teach} {LLMs} to {Ask} {Clarifying} {Questions}},
 url = {https://arxiv.org/abs/2410.13788},
 volume = {abs/2410.13788},
 year = {2024}
}

@misc{malaviya_contextualized_2024,
 author = {Malaviya, Chaitanya and Chang, Joseph Chee and Roth, Dan and Iyyer, Mohit and Yatskar, Mark and Lo, Kyle},
 journal = {ArXiv preprint},
 title = {Contextualized {Evaluations}: {Taking} the {Guesswork} {Out} of {Language} {Model} {Evaluations}},
 url = {https://arxiv.org/abs/2411.07237},
 volume = {abs/2411.07237},
 year = {2024}
}

@misc{kuhn_clam_2023,
 author = {Kuhn, Lorenz and Gal, Yarin and Farquhar, Sebastian},
 journal = {ArXiv preprint},
 title = {{CLAM}: {Selective} {Clarification} for {Ambiguous} {Questions} with {Generative} {Language} {Models}},
 url = {https://arxiv.org/abs/2212.07769},
 volume = {abs/2212.07769},
 year = {2022}
}

@misc{li_eliciting_2023,
 author = {Li, Belinda Z. and Tamkin, Alex and Goodman, Noah and Andreas, Jacob},
 journal = {ArXiv preprint},
 title = {Eliciting {Human} {Preferences} with {Language} {Models}},
 url = {https://arxiv.org/abs/2310.11589},
 volume = {abs/2310.11589},
 year = {2023}
}

@article{xie2024osworld,
 author = {Xie, Tianbao and Zhang, Danyang and Chen, Jixuan and Li, Xiaochuan and Zhao, Siheng and Cao, Ruisheng and Hua, Toh J and Cheng, Zhoujun and Shin, Dongchan and Lei, Fangyu and others},
 journal = {Advances in Neural Information Processing Systems},
 pages = {52040--52094},
 title = {Osworld: Benchmarking multimodal agents for open-ended tasks in real computer environments},
 volume = {37},
 year = {2024}
}

@article{chen2025decisionflow,
 author = {Chen, Xiusi and Wang, Shanyong and Qian, Cheng and Wang, Hongru and Han, Peixuan and Ji, Heng},
 journal = {ArXiv preprint},
 title = {DecisionFlow: Advancing Large Language Model as Principled Decision Maker},
 url = {https://arxiv.org/abs/2505.21397},
 volume = {abs/2505.21397},
 year = {2025}
}

@article{zhou2023webarena,
 author = {Zhou, Shuyan and Xu, Frank F and Zhu, Hao and Zhou, Xuhui and Lo, Robert and Sridhar, Abishek and Cheng, Xianyi and Ou, Tianyue and Bisk, Yonatan and Fried, Daniel and others},
 journal = {ArXiv preprint},
 title = {Webarena: A realistic web environment for building autonomous agents},
 url = {https://arxiv.org/abs/2307.13854},
 volume = {abs/2307.13854},
 year = {2023}
}

@article{hu2024uncertainty,
 author = {Hu, Zhiyuan and Liu, Chumin and Feng, Xidong and Zhao, Yilun and Ng, See-Kiong and Luu, Anh Tuan and He, Junxian and Koh, Pang Wei and Hooi, Bryan},
 journal = {ArXiv preprint},
 title = {Uncertainty of thoughts: Uncertainty-aware planning enhances information seeking in large language models},
 url = {https://arxiv.org/abs/2402.03271},
 volume = {abs/2402.03271},
 year = {2024}
}

@inproceedings{tian2023just,
 address = {Singapore},
 author = {Tian, Katherine  and
Mitchell, Eric  and
Zhou, Allan  and
Sharma, Archit  and
Rafailov, Rafael  and
Yao, Huaxiu  and
Finn, Chelsea  and
Manning, Christopher},
 booktitle = {Proceedings of the 2023 Conference on Empirical Methods in Natural Language Processing},
 doi = {10.18653/v1/2023.emnlp-main.330},
 editor = {Bouamor, Houda  and
Pino, Juan  and
Bali, Kalika},
 pages = {5433--5442},
 publisher = {Association for Computational Linguistics},
 title = {Just Ask for Calibration: Strategies for Eliciting Calibrated Confidence Scores from Language Models Fine-Tuned with Human Feedback},
 url = {https://aclanthology.org/2023.emnlp-main.330},
 year = {2023}
}

@article{dong2024can,
 author = {Dong, Yijiang River and Hu, Tiancheng and Collier, Nigel},
 journal = {ArXiv preprint},
 title = {Can LLM be a Personalized Judge?},
 url = {https://arxiv.org/abs/2406.11657},
 volume = {abs/2406.11657},
 year = {2024}
}

@article{lin2022inferring,
  title={Inferring rewards from language in context},
  author={Lin, Jessy and Fried, Daniel and Klein, Dan and Dragan, Anca},
  journal={arXiv preprint arXiv:2204.02515},
  year={2022}
}

@article{qiu2025bayesian,
  title={Bayesian teaching enables probabilistic reasoning in large language models},
  author={Qiu, Linlu and Sha, Fei and Allen, Kelsey and Kim, Yoon and Linzen, Tal and van Steenkiste, Sjoerd},
  journal={arXiv preprint arXiv:2503.17523},
  year={2025}
}

@article{ren2023robots,
  title={Robots that ask for help: Uncertainty alignment for large language model planners},
  author={Ren, Allen Z and Dixit, Anushri and Bodrova, Alexandra and Singh, Sumeet and Tu, Stephen and Brown, Noah and Xu, Peng and Takayama, Leila and Xia, Fei and Varley, Jake and others},
  journal={arXiv preprint arXiv:2307.01928},
  year={2023}
}

@article{wu2025collabllm,
 author = {Wu, Shirley and Galley, Michel and Peng, Baolin and Cheng, Hao and Li, Gavin and Dou, Yao and Cai, Weixin and Zou, James and Leskovec, Jure and Gao, Jianfeng},
 journal = {ArXiv preprint},
 title = {CollabLLM: From Passive Responders to Active Collaborators},
 url = {https://arxiv.org/abs/2502.00640},
 volume = {abs/2502.00640},
 year = {2025}
}

@article{frank2012predicting,
  title={Predicting pragmatic reasoning in language games},
  author={Frank, Michael C and Goodman, Noah D},
  journal={Science},
  volume={336},
  number={6084},
  pages={998--998},
  year={2012},
  publisher={American Association for the Advancement of Science}
}

@article{goodman2016pragmatic,
  title={Pragmatic language interpretation as probabilistic inference},
  author={Goodman, Noah D and Frank, Michael C},
  journal={Trends in cognitive sciences},
  volume={20},
  number={11},
  pages={818--829},
  year={2016},
  publisher={Elsevier}
}

@article{grand2025shoot,
  title={Shoot First, Ask Questions Later? Building Rational Agents that Explore and Act Like People},
  author={Grand, Gabriel and Pepe, Valerio and Andreas, Jacob and Tenenbaum, Joshua B},
  journal={arXiv preprint arXiv:2510.20886},
  year={2025}
}

@inproceedings{hawkins2015you,
  title={Why do you ask? Good questions provoke informative answers},
  author={Hawkins, Robert XD and Stuhlmuller, Andreas and Degen, Judith and Goodman, Noah D},
  booktitle={Proceedings of the Annual Meeting of the Cognitive Science Society},
  volume={37},
  year={2015}
}

@misc{zhou2025usereffective,
  author = {Zhou, Xuhui and Sun, Weiwei},
  title = {The Quest of {User}-Effective {AI} Agents},
  year = {2025},
  url = {https://xuhuizhou.github.io/blog/on-the-quest-of-user-effective-ai-agents},
  note = {Blog post. Accessed: 2026-01-06},
}

@article{sun2025training,
  title={Training Proactive and Personalized LLM Agents},
  author={Sun, Weiwei and Zhou, Xuhui and Du, Weihua and Wang, Xingyao and Welleck, Sean and Neubig, Graham and Sap, Maarten and Yang, Yiming},
  journal={arXiv preprint arXiv:2511.02208},
  year={2025}
}

@article{qian2025userrl,
  title={UserRL: Training Interactive User-Centric Agent via Reinforcement Learning},
  author={Qian, Cheng and Liu, Zuxin and Prabhakar, Akshara and Qiu, Jielin and Liu, Zhiwei and Chen, Haolin and Kokane, Shirley and Ji, Heng and Yao, Weiran and Heinecke, Shelby and others},
  journal={arXiv preprint arXiv:2509.19736},
  year={2025}
}

@article{monroe-etal-2017-colors,
    title = "Colors in Context: A Pragmatic Neural Model for Grounded Language Understanding",
    author = "Monroe, Will  and
      Hawkins, Robert X.D.  and
      Goodman, Noah D.  and
      Potts, Christopher",
    editor = "Lee, Lillian  and
      Johnson, Mark  and
      Toutanova, Kristina",
    journal = "Transactions of the Association for Computational Linguistics",
    volume = "5",
    year = "2017",
    address = "Cambridge, MA",
    publisher = "MIT Press",
    url = "https://aclanthology.org/Q17-1023/",
    doi = "10.1162/tacl_a_00064",
    pages = "325--338"
}

@misc{cao2025pragmaticreasoningimprovesllm,
      title={Pragmatic Reasoning improves LLM Code Generation}, 
      author={Zhuchen Cao and Sven Apel and Adish Singla and Vera Demberg},
      year={2025},
      eprint={2502.15835},
      archivePrefix={arXiv},
      primaryClass={cs.CL},
      url={https://arxiv.org/abs/2502.15835}, 
}

@inproceedings{wang-demberg-2024-rsa,
    title = "{RSA}-Control: A Pragmatics-Grounded Lightweight Controllable Text Generation Framework",
    author = "Wang, Yifan  and
      Demberg, Vera",
    editor = "Al-Onaizan, Yaser  and
      Bansal, Mohit  and
      Chen, Yun-Nung",
    booktitle = "Proceedings of the 2024 Conference on Empirical Methods in Natural Language Processing",
    month = nov,
    year = "2024",
    address = "Miami, Florida, USA",
    publisher = "Association for Computational Linguistics",
    url = "https://aclanthology.org/2024.emnlp-main.318/",
    doi = "10.18653/v1/2024.emnlp-main.318",
    pages = "5561--5582"
}

@inproceedings{andreas-klein-2016-reasoning,
    title = "Reasoning about Pragmatics with Neural Listeners and Speakers",
    author = "Andreas, Jacob  and
      Klein, Dan",
    editor = "Su, Jian  and
      Duh, Kevin  and
      Carreras, Xavier",
    booktitle = "Proceedings of the 2016 Conference on Empirical Methods in Natural Language Processing",
    month = nov,
    year = "2016",
    address = "Austin, Texas",
    publisher = "Association for Computational Linguistics",
    url = "https://aclanthology.org/D16-1125/",
    doi = "10.18653/v1/D16-1125",
    pages = "1173--1182"
}

@article{yao2024tau,
  title={tau-bench: A Benchmark for Tool-Agent-User Interaction in Real-World Domains},
  author={Yao, Shunyu and Shinn, Noah and Razavi, Pedram and Narasimhan, Karthik},
  journal={arXiv preprint arXiv:2406.12045},
  year={2024}
}

@article{sumers2021extending,
  title={Extending rational models of communication from beliefs to actions},
  author={Sumers, Theodore R and Hawkins, Robert D and Ho, Mark K and Griffiths, Thomas L},
  journal={arXiv preprint arXiv:2105.11950},
  year={2021}
}

@inproceedings{
yao2023react,
title={ReAct: Synergizing Reasoning and Acting in Language Models},
author={Shunyu Yao and Jeffrey Zhao and Dian Yu and Nan Du and Izhak Shafran and Karthik R Narasimhan and Yuan Cao},
booktitle={The Eleventh International Conference on Learning Representations },
year={2023},
url={https://openreview.net/forum?id=WE_vluYUL-X}
}

@ARTICLE{4082064,
  author={Howard, Ronald A.},
  journal={IEEE Transactions on Systems Science and Cybernetics}, 
  title={Information Value Theory}, 
  year={1966},
  volume={2},
  number={1},
  pages={22-26},
  doi={10.1109/TSSC.1966.300074}}

@book{raiffa1961applied,
  title={Applied Statistical Decision Theory},
  author={Raiffa, H. and Schlaifer, R.},
  series={Studies in managerial economics},
  url={https://books.google.co.uk/books?id=SpO0KFcFQDsC},
  year={1961},
  publisher={Division of Research, Graduate School of Business Administration, Harvard University}
}

@inproceedings{peng2024plga,
  author       = {Andi Peng and
                  Andreea Bobu and
                  Belinda Z. Li and
                  Theodore R. Sumers and
                  Ilia Sucholutsky and
                  Nishanth Kumar and
                  Thomas L. Griffiths and
                  Julie A. Shah},
  editor       = {Dan Grollman and
                  Elizabeth Broadbent and
                  Wendy Ju and
                  Harold Soh and
                  Tom Williams},
  title        = {Preference-Conditioned Language-Guided Abstraction},
  booktitle    = {Proceedings of the 2024 {ACM/IEEE} International Conference on Human-Robot
                  Interaction, {HRI} 2024, Boulder, CO, USA, March 11-15, 2024},
  pages        = {572--581},
  publisher    = {{ACM}},
  year         = {2024},
  url          = {https://doi.org/10.1145/3610977.3634930},
  doi          = {10.1145/3610977.3634930},
  bibsource    = {dblp computer science bibliography, https://dblp.org}
}

@article{hu2025simbench,
  title={Simbench: Benchmarking the ability of large language models to simulate human behaviors},
  author={Hu, Tiancheng and Baumann, Joachim and Lupo, Lorenzo and Collier, Nigel and Hovy, Dirk and R{\"o}ttger, Paul},
  journal={arXiv preprint arXiv:2510.17516},
  year={2025}
}

@article{chen2026decoupling,
  title={Decoupling the Effect of Chain-of-Thought Reasoning: A Human Label Variation Perspective},
  author={Chen, Beiduo and Hu, Tiancheng and Zhang, Caiqi and Litschko, Robert and Korhonen, Anna and Plank, Barbara},
  journal={arXiv preprint arXiv:2601.03154},
  year={2026}
}

@article{zhang2026confidence,
  title={Confidence Estimation for LLMs in Multi-turn Interactions},
  author={Zhang, Caiqi and Yang, Ruihan and Zhu, Xiaochen and Li, Chengzu and Hu, Tiancheng and Dong, Yijiang River and Yang, Deqing and Collier, Nigel},
  journal={arXiv preprint arXiv:2601.02179},
  year={2026}
}

@article{hu2025navigating,
  title={Navigating the Alignment-Calibration Trade-off: A Pareto-Superior Frontier via Model Merging},
  author={Hu, Tiancheng and Minixhofer, Benjamin and Collier, Nigel},
  journal={arXiv preprint arXiv:2510.17426},
  year={2025}
}

@inproceedings{dong-etal-2025-personalization,
    title = "When Personalization Meets Reality: A Multi-Faceted Analysis of Personalized Preference Learning",
    author = {Dong, Yijiang River  and
      Hu, Tiancheng  and
      Liu, Yinhong  and
      {\"U}st{\"u}n, Ahmet  and
      Collier, Nigel},
    editor = "Christodoulopoulos, Christos  and
      Chakraborty, Tanmoy  and
      Rose, Carolyn  and
      Peng, Violet",
    booktitle = "Findings of the Association for Computational Linguistics: EMNLP 2025",
    month = nov,
    year = "2025",
    address = "Suzhou, China",
    publisher = "Association for Computational Linguistics",
    url = "https://aclanthology.org/2025.findings-emnlp.916/",
    doi = "10.18653/v1/2025.findings-emnlp.916",
    pages = "16880--16894",
    ISBN = "979-8-89176-335-7"}

\newpage
\appendix

\begin{figure*}[ht!]
    \includegraphics[width=\linewidth]{figures/case_study.pdf}
    \caption{A side by side comparison for different methods for Mixed 20 Question task. The figure contrasts four controllers---No-Ask, Fixed-Round, Confidence Thresholding ($\tau=0.90$), and our VOI policy---on a single Mixed 20Q instance with communication cost $c=0.05$. Task stakes are encoded directly in the terminal utility: a correct animal guess yields reward $1$ (low stakes), whereas a correct medical diagnosis yields reward $10$ (high stakes). The objective maximizes decision utility minus dialogue cost, $U(\theta, a) - c(\xi)$.}
    \label{fig: case study}
\end{figure*}

\section{Case Study: VoI is Risk-Aware}
Figure~\ref{fig: case study} provides a compelling qualitative example of why the VoI framework is superior to heuristic-based methods like confidence thresholding. The experiment contrasts a low-stakes task (guessing an animal, reward=1) with a high-stakes task (medical diagnosis, reward=10), using an identical communication cost ($c=0.05$).

In the \textbf{high-stakes medical diagnosis} (Fig.~\ref{fig: case study}b), the potential reward for a correct answer is high. The VoI agent correctly calculates that even questions with moderate information gain are valuable enough to outweigh the communication cost. It, therefore, continues to ask clarifying questions until it is highly confident, stopping several rounds \emph{after} the confidence-thresholding baseline would have stopped, even though significant ambiguity remains, leading to an incorrect diagnosis.

In the \textbf{low-stakes animal guessing game} (Fig.~\ref{fig: case study}a), the maximum potential utility is low. Here, the VoI agent correctly assesses that the potential utility gain from asking many questions is not worth the cumulative communication cost. It, therefore, halts the conversation earlier than the confidence-thresholding method, avoiding unnecessary cognitive load on the user for a low-risk task. The confidence-based agent, blind to the low stakes, would have continued asking questions, needlessly imposing cognitive load on the user for a trivial task.

This case study reveals that effective communication requires balancing two distinct pressures: the drive to reduce uncertainty (an epistemic goal) and the need to consider the task's stakes (a utilitarian goal). Confidence-based methods address only the former. The VoI framework excels because it naturally unifies both: it quantifies the value of reducing uncertainty precisely in terms of its expected impact on the final, stake-weighted utility. This principled balance enables the agent to be appropriately cautious in high-stakes scenarios and efficient in low-stakes ones—a critical capability for building trustworthy and effective human-AI collaborators.

\section{Main Results in Tables}
\begin{figure*}[ht]
\footnotesize
\centering
\caption{GPT-4: results for different methods and thresholds across three tasks. For \textbf{Webshop}, LLM is normalized by 10 and utilities are $\text{Util}=\text{LLM}-\#T \times \{0.01,0.05\}$. Mixed 20Q utilities are recomputed per spec. Within each \textit{method}, the best utility is \underline{underlined}. The global best per task/cost is \textbf{\textit{bold+italic}} and the second best is \textbf{bold}.}
\resizebox{\textwidth}{!}{
\begin{tabular}{l|*{7}{c}|*{5}{c}|*{5}{c}}
\toprule
\multirow{2}{*}{\bf Method}
& \multicolumn{7}{c}{\bf Mixed 20Q}
& \multicolumn{5}{|c|}{\bf Flight Rec.}
& \multicolumn{5}{|c|}{\bf Webshop} \\
\cmidrule(lr){2-8}\cmidrule(lr){9-13}\cmidrule(lr){14-18}
& \makecell[t]{$\boldsymbol{\tau}$} & \makecell[t]{\textbf{Acc.}\\\textbf{(Animal)}} & \makecell[t]{\textbf{Acc.}\\\textbf{(Med)}} & \makecell[t]{\textbf{\#T}\\\textbf{(Animal)}} & \makecell[t]{\textbf{\#T}\\\textbf{(Med)}} & \makecell[t]{\textbf{Util.}\\\textbf{(0.01)}} & \makecell[t]{\textbf{Util.}\\\textbf{(0.05)}} &
\makecell[t]{$\boldsymbol{\tau}$} & \makecell[t]{\textbf{Reward}} & \makecell[t]{\textbf{\#T}}
& \makecell[t]{\textbf{Util.}\\\textbf{(0.01)}}
& \makecell[t]{\textbf{Util.}\\\textbf{(0.05)}} &
\makecell[t]{$\boldsymbol{\tau}$} & \makecell[t]{\textbf{LLM}} & \makecell[t]{\textbf{\#T}} & \makecell[t]{\textbf{Util.}\\\textbf{(0.01)}} & \makecell[t]{\textbf{Util.}\\\textbf{(0.05)}} \\
\midrule
No Question
& -- & 0.01 & 0.06 & 0.00 & 0.00 & \underline{0.70} & \underline{0.70} & -- & 0.17 & 0.00 & \underline{0.17} & \underline{0.17} & -- & 0.54 & 0.00 & \underline{0.54} & \underline{\textbf{\textit{0.54}}} \\
Adaptive 
& -- & 0.68 & 0.53 & 17.80 & 6.254 & \underline{10.26} & \underline{2.89} & -- & 0.20 & 0.56 & \underline{0.20} & \underline{0.17} & -- & 0.57 & 0.89 & \underline{0.56} & \underline{0.52} \\
\cdashline{1-18}\noalign{\vskip 1pt}
\multirow[t]{4}{*}{Fixed Round}
  & 5  & 0.24 & 0.51 & 5.00 & 5.00 & 6.95 & 4.75 & 1.00 & 0.22 & 1.00 & 0.21 & 0.17 & 1.00 & 0.56 & 1.00 & 0.55 & 0.51 \\
& 10 & 0.60 & 0.78 & 10.00 & 10.00 & 12.70 & \underline{8.30} & 2.00 & 0.32 & 2.00 & \textbf{0.30} & \underline{0.22} & 2.00 & 0.57 & 2.00 & 0.55 & 0.47 \\
& 15 & 0.77 & 0.78 & 15.00 & 10.00 & 13.90 & 7.50 & 3.00 & 0.35 & 3.00 & \underline{\textbf{\textit{0.32}}} & 0.20 & 3.00 & 0.62 & 3.00 & \underline{0.59} & 0.47 \\
& 20 & 0.87 & 0.78 & 20.00 & 10.00 & \underline{\textbf{\textit{14.40}}} & 6.00 & 4.00 & 0.36 & 4.00 & \underline{\textbf{\textit{0.32}}} & 0.16 & 4.00 & 0.63 & 4.00 & \underline{0.59} & 0.43 \\
\cdashline{1-18}\noalign{\vskip 1pt}
\multirow[t]{3}{*}{Confidence}
  & 0.50 & 0.20 & 0.31 & 4.01 & 2.54 & 4.67 & 2.97 & 0.50 & 0.19 & 0.71 & 0.19 & 0.16 & 0.50 & 0.55 & 0.78 & 0.54 & 0.51 \\
& 0.70 & 0.45 & 0.60 & 5.68 & 4.56 & 9.89 & 7.43 & 0.70 & 0.23 & 1.09 & \underline{0.22} & \underline{0.17} & 0.70 & 0.60 & 1.31 & 0.58 & \underline{\textbf{0.53}} \\
& 0.90 & 0.59 & 0.65 & 8.48 & 6.49 & \underline{11.49} & \underline{7.84} & 0.90 & 0.24 & 2.82 & \underline{0.21} & 0.10 & 0.90 & 0.63 & 2.95 & \underline{\textbf{\textit{0.60}}} & 0.48 \\
\cdashline{1-18}\noalign{\vskip 1pt}
\multirow[t]{2}{*}{VOI}
  & 0.01 & 0.76 & 0.78 & 11.80 & 8.07 & \underline{\textbf{14.14}} & \underline{\textbf{\textit{9.10}}} & 0.01 & 0.36 & 1.49 & \textbf{0.35} & \underline{\textbf{\textit{0.28}}} & 0.01 & 0.63 & 2.95 & \underline{\textbf{\textit{0.60}}} & 0.49 \\
& 0.05 & 0.74 & 0.78 & 11.46 & 7.99 & 13.97 & \textbf{9.07} & 0.05 & 0.29 & 0.82 & \underline{0.29} & \textbf{0.25} & 0.05 & 0.61 & 1.74 & 0.59 & \underline{0.52} \\
\bottomrule
\end{tabular}}
\label{fig:prompting-gpt4}
\end{figure*}

\begin{table*}[!ht]
\footnotesize
\centering
\caption{Gemini-2.5-Flash: results for different methods and thresholds across three tasks. Format is the same as Figure \ref{fig:prompting-gpt4}}
\resizebox{\textwidth}{!}{
\begin{tabular}{l|ccccccc|ccccc|ccccc}
\toprule
\multirow{2}{*}{\bf Method} &
\multicolumn{7}{c|}{\bf Mixed 20Q} &
\multicolumn{5}{c|}{\bf Flight Rec.} &
\multicolumn{5}{c}{\bf Webshop} \\
\cmidrule(lr){2-8}\cmidrule(lr){9-13}\cmidrule(lr){14-18}
 & \makecell[t]{$\boldsymbol{\tau}$}
 & \makecell[t]{\textbf{Acc.}\\\textbf{(Animal)}}
 & \makecell[t]{\textbf{Acc.}\\\textbf{(Med)}}
 & \makecell[t]{\textbf{\#T}\\\textbf{(Animal)}}
 & \makecell[t]{\textbf{\#T}\\\textbf{(Med)}}
 & \makecell[t]{\textbf{Util.}\\\textbf{(0.01)}}
 & \makecell[t]{\textbf{Util.}\\\textbf{(0.05)}}
 & \makecell[t]{$\boldsymbol{\tau}$}
 & \makecell[t]{\textbf{Reward}}
 & \makecell[t]{\textbf{\#T}}
 & \makecell[t]{\textbf{Util.}\\\textbf{(0.01)}}
 & \makecell[t]{\textbf{Util.}\\\textbf{(0.05)}}
 & \makecell[t]{$\boldsymbol{\tau}$}
 & \makecell[t]{\textbf{LLM}}
 & \makecell[t]{\textbf{\#T}}
 & \makecell[t]{\textbf{Util.}\\\textbf{(0.01)}}
 & \makecell[t]{\textbf{Util.}\\\textbf{(0.05)}} \\
\midrule
No Question
 & --  & 0.01 & 0.06 & 0.00 & 0.00 & \underline{0.70} & \textbf{\underline{0.70}}
 & --  & 0.16 & 0.00 & \underline{0.16} & \underline{0.16}
 & --  & 0.50 & 0.00 & \underline{0.50} & \textbf{\underline{0.50}} \\
Adaptive
 & --  & 0.28 & 0.37 & 4.78 & 6.36 & \underline{5.96} & \textbf{\underline{3.79}}
 & --  & 0.22 & 0.21 & \underline{0.22} & \underline{0.21}
 & --  & 0.51 & 0.55 & \underline{0.51} & \underline{0.48} \\
\cdashline{1-18}\noalign{\vskip 1pt}
Fixed Round
 & 5   & 0.16 & 0.29 & 5.00 & 5.00 & 3.95 & \underline{1.75}
 & 1.00 & 0.18 & 1.00 & \underline{0.17} & \underline{0.13}
 & 1.00 & 0.55 & 1.00 & 0.54 & \textbf{\underline{0.50}} \\
 & 10  & 0.33 & 0.30 & 10.00 & 10.00 & 5.20 & 0.80
 & 2.00 & 0.18 & 2.00 & 0.16 & 0.08
 & 2.00 & 0.57 & 2.00 & 0.55 & 0.47 \\
 & 15  & 0.40 & 0.30 & 15.00 & 10.00 & \underline{5.40} & -1.00
 & 3.00 & 0.19 & 3.00 & 0.16 & 0.04
 & 3.00 & 0.59 & 3.00 & \textbf{0.56} & 0.44 \\
 & 20  & 0.39 & 0.30 & 20.00 & 10.00 & 4.80 & -3.60
 & 4.00 & 0.21 & 4.00 & \underline{0.17} & 0.01
 & 4.00 & 0.61 & 4.00 & \textit{\textbf{\underline{0.57}}} & 0.41 \\
\cdashline{1-18}\noalign{\vskip 1pt}
Confidence
 & 0.50 & 0.22 & 0.27 & 4.87 & 5.12 & 4.36 & \underline{2.21}
 & 0.50 & 0.14 & 0.09 & 0.14 & 0.14
 & 0.50 & 0.52 & 0.48 & 0.52 & 0.50 \\
 & 0.70 & 0.16 & 0.31 & 5.06 & 6.08 & 4.13 & 1.87
 & 0.70 & 0.20 & 0.99 & 0.19 & 0.15
 & 0.70 & 0.54 & 0.55 & 0.54 & \textit{\textbf{\underline{0.51}}} \\
 & 0.90 & 0.36 & 0.30 & 11.28 & 9.25 & \underline{5.38} & 0.50
 & 0.90 & 0.25 & 1.53 & \underline{0.24} & \underline{0.17}
 & 0.90 & 0.59 & 2.73 & \textbf{\underline{0.56}} & 0.45 \\
\cdashline{1-18}\noalign{\vskip 1pt}
VOI
 & 0.01 & 0.28 & 0.55 & 8.48 & 7.63 & \textit{\textbf{\underline{7.38}}} & 3.68
 & 0.01 & 0.30 & 1.62 & \textit{\textbf{\underline{0.28}}} & \textbf{0.22}
 & 0.01 & 0.59 & 2.15 & \textit{\textbf{\underline{0.57}}} & 0.48 \\
 & 0.05 & 0.15 & 0.50 & 4.20 & 6.99 & \textbf{6.01} & \textit{\textbf{\underline{4.05}}}
 & 0.05 & 0.28 & 1.07 & \textbf{0.27} & \textit{\textbf{\underline{0.23}}}
 & 0.05 & 0.56 & 1.20 & 0.55 & \textbf{\underline{0.50}} \\
\bottomrule
\end{tabular}}
\end{table*}

\newpage
\section{Prompts}
\label{app: prompts}
\onecolumn
\subsection{Mixed 20 Questions}

\begin{figure*}[htbp]
\begin{tcolorbox}[title={Animal — Direct Prompting}]
You are playing 20 Questions as the guesser. Your goal is to figure out what animal I'm thinking of by asking questions.\\
You have asked \{question\_count\} questions so far and have \{remaining\_questions\} questions left.\\
The possible animals you're trying to guess include: \{answer\_set\}\\
\textbf{Rules:}\\
1. Ask only yes/no questions (answerable with ``Yes''/``No'').\\
2. Ask one question at a time.\\
3. Keep asking until you use all 20 questions.\\
4. Do not ask the same question twice.\\
5. Do not guess a specific animal early (e.g., ``Is it a cat?''). Start broad to narrow options.
\end{tcolorbox}
\caption{Direct Prompting (Animal 20 Question)}
\label{fig:animal_direct_prompting}
\end{figure*}

\begin{figure*}[htbp]
\begin{tcolorbox}[title={Animal — Auto Stop}]
You are playing 20 Questions as the guesser. Your goal is to figure out what animal I'm thinking of.\\
You have asked \{question\_count\} questions so far and have \{remaining\_questions\} questions left.\\
The possible animals you're trying to guess include: \{answer\_set\}\\
\textbf{Rules:}\\
1. Ask only yes/no questions.\\
2. Ask one question at a time.\\
3. When you're ready to guess, output: \texttt{My guess is: [animal]}.\\
4. Do not ask the same question twice.\\
5. Avoid premature specific guesses; use broader categories to narrow down.
\end{tcolorbox}
\caption{Auto Stop (Animal 20 Question)}
\label{fig:animal_auto_stop}
\end{figure*}

\begin{figure*}[htbp]
\begin{tcolorbox}[title={Animal — Confidence Thresholding}]
You are tasked with guessing the animal based on the entire conversation. Provide a final guess, even if uncertain. The set of possible animals is: \{answer\_set\}.\\
Also estimate your confidence from 1 to 100 (100 = highest).\\
\textbf{Output exactly:} \texttt{\{\,"guess": "\textless animal\textgreater", "confidence": \textless number 1--100\textgreater\,\}}\\
\textbf{Confidence Scale (1--100):}\\
1--20: Very minimal evidence.\\
21--40: Weak evidence.\\
41--60: Moderate evidence.\\
61--80: Strong evidence.\\
81--100: Very strong evidence.
\end{tcolorbox}
\caption{Confidence Thresholding (Animal 20 Question)}
\label{fig:animal_conf_threshold}
\end{figure*}

\begin{figure*}[htbp]
\begin{tcolorbox}[title={Animal — VOI: Question Generation}]
I'm playing a game of 20 Questions to identify an animal. Based on previous questions and answers:\\
\{previous\_qa\}\\
Generate 5 yes/no questions that help identify the animal.\\
Format as a numbered list.
\end{tcolorbox}
\caption{VOI: Question Generation (Animal 20 Question)}
\label{fig:animal_voi_qgen}
\end{figure*}

\begin{figure*}[htbp]
\begin{tcolorbox}[title={Animal — VOI: Batch Answer Simulation}]
You are playing a game of Twenty Questions. You will receive a yes/no question and a list of animals.\\
Answer for each animal with: \textbf{Yes} (clearly correct), \textbf{No} (clearly incorrect), or \textbf{Maybe} (uncertain/ambiguous/varies by species or not confident). Prefer ``Maybe'' over guessing.\\
\textbf{Format exactly as:} \texttt{Animal Name: Yes/No/Maybe}\\
\textbf{Question:} \{question\}\\
\textbf{Animals:} \{candidate\_list\}
\end{tcolorbox}
\caption{VOI: Batch Answer Simulation (Animal 20 Question)}
\label{fig:animal_voi_batch}
\end{figure*}


\begin{figure*}[htbp]
\begin{tcolorbox}[title={Medical — Direct Prompting}]
You are the doctor and must diagnose the patient using only yes/no questions.\\
You have asked \{question\_count\} questions so far and have \{remaining\_questions\} left.\\
Possible diagnoses: \{answer\_set\}\\
You may ask up to 20 yes/no questions to understand the condition. At the end, output your diagnosis.
\end{tcolorbox}
\caption{Direct Prompting (Medical Diagnosis)}
\label{fig:medical_direct_prompting}
\end{figure*}

\begin{figure*}[htbp]
\begin{tcolorbox}[title={Medical — Auto Stop}]
You are the doctor and may ask up to 20 yes/no questions to diagnose the patient.\\
You have asked \{question\_count\} questions so far and have \{remaining\_questions\} left.\\
Possible diagnoses: \{answer\_set\}\\
You can ask up to 10 yes/no questions. Stop when you have enough information.\\
\textbf{Format your guess as:} \texttt{My guess is: [diagnosis]}.
\end{tcolorbox}
\caption{Auto Stop (Medical Diagnosis)}
\label{fig:medical_auto_stop}
\end{figure*}

\begin{figure*}[htbp]
\begin{tcolorbox}[title={Medical — Confidence Thresholding}]
Diagnose the patient based on the entire conversation. Provide a final diagnosis, even if uncertain. Set of diseases: \{answer\_set\}.\\
Also estimate your confidence (1--100).\\
\textbf{Output exactly:} \texttt{\{\,"guess": "\textless diagnosis\textgreater", "confidence": \textless number 1--100\textgreater\,\}}\\
\textbf{Confidence Scale (1--100):}\\
1--20: Extremely uncertain.\\
21--40: Weak evidence.\\
41--60: Moderate evidence.\\
61--80: Strong evidence.\\
81--100: Very strong evidence.
\end{tcolorbox}
\caption{Confidence Thresholding (Medical Diagnosis)}
\label{fig:medical_conf_threshold}
\end{figure*}

\begin{figure*}[htbp]
\begin{tcolorbox}[title={Medical — VOI: Question Generation}]
I'm a doctor trying to diagnose a patient's condition through a series of questions. Based on symptoms and previous answers:\\
\{previous\_qa\}\\
Generate 5 yes/no questions that most effectively narrow the possible conditions (roughly halving the set each time).\\
Focus on distinguishing symptoms, risk factors, or medical history.\\
Format as a numbered list.
\end{tcolorbox}
\caption{VOI: Question Generation (Medical Diagnosis)}
\label{fig:medical_voi_qgen}
\end{figure*}

\begin{figure*}[htbp]
\begin{tcolorbox}[title={Medical — VOI: Batch Answer Simulation}]
I'm a medical diagnostician. Below is a yes/no question and a list of medical conditions.\\
\textbf{Question:} ``\{question\}''\\
For each condition, answer with just ``Yes'' or ``No'', based on typical presentation.\\
\textbf{Reply exactly as:} \texttt{Condition: Answer}\\
\textbf{Conditions:} \{candidate\_list\}
\end{tcolorbox}
\caption{VOI: Batch Answer Simulation (Medical Diagnosis)}
\label{fig:medical_voi_batch}
\end{figure*}

\newpage

\subsection{Flight Recommendation}

\begin{figure*}[htbp]
\begin{tcolorbox}[title={Direct Prompting and Confidence Thresholding}]
\textbf{Opening}\\
User: Help me pick flights. My preferences are fixed; infer them and choose. Use your best judgement; don't ask for more info.\\[2pt]
\textbf{--- SUPPORT HISTORY ---}\\
User: Which flight is best?\\
Flight 1: \{option 1\}\\
Flight 2: \{option 2\}\\
Flight 3: \{option 3\}\\
User: I prefer flight \{1/2/3\}\\[2pt]
\textbf{NEW Round (no answer shown)}\\
User: Which flight is best?\\
Flight 1: \{option 1\}\\
Flight 2: \{option 2\}\\
Flight 3: \{option 3\}\\[2pt]
\textbf{Required Output}\\
Model: The best option is Flight
\end{tcolorbox}
\caption{The prompt used for Direct Prompting and Confidence Thresholding. Logit is extracted as measure of confidence.}
\end{figure*}

\begin{figure*}[htbp]
\begin{tcolorbox}[title={VOI — Prior over Feature States}]
You are calibrating a probabilistic user model.\\

Feature: \{feature\}\\
History (support + any clarifying Q\&A):\\
\{history\_ctx\}\\[2pt]

Based \emph{only} on this history, estimate $P(\text{state})$ for the feature.\\
\textbf{Return STRICT JSON with keys exactly} \{\{states\}\} \textbf{that sum to 1}. Example: \texttt{\{"lower": 0.33, "higher": 0.33, "none": 0.34\}}\\
\textbf{JSON:}
\end{tcolorbox}
\caption{Prior Estimation for VOI (Airline Preference Matching)}
\label{fig:voi_prior}
\end{figure*}

\begin{figure*}[htbp]
\begin{tcolorbox}[title={VOI — Posterior with Options}]
You are calibrating a probabilistic user model.\\

Feature: \{feature\}\\
History (support + any clarifying Q\&A):\\
\{history\_ctx\}\\[2pt]

Current options:\\
A) \{option A\}\\
B) \{option B\}\\
C) \{option C\}\\[2pt]

Estimate the \emph{posterior} distribution over the user's \{feature\} state given full history.\\
\textbf{Return STRICT JSON with keys exactly} \{\{states\}\} \textbf{that sum to 1}.\\
\textbf{JSON:}
\end{tcolorbox}
\caption{Posterior Estimation with Options (Airline Preference Matching)}
\label{fig:voi_posterior}
\end{figure*}

\begin{figure*}[htbp]
\begin{tcolorbox}[title={VOI — Candidate Preference Questions}]
You are an AI assistant helping a user choose between flight options A, B, and C. You've analyzed the support examples but still have some uncertainty.\\[2pt]

\{support\_history\}\\
\{qa\_context\}\\[4pt]

Generate one multiple-choice question about a single aspect of the user's preference that will help decide among the options below.\\[2pt]

A) \{option A\}\\
B) \{option B\}\\
C) \{option C\}\\[2pt]

\textbf{Question:}
\end{tcolorbox}
\caption{VOI Candidate Questions (Airline Preference Matching)}
\label{fig:voi_candidates}
\end{figure*}

\end{document}